\pdfoutput=1
\documentclass[11pt]{article}

\usepackage[preprint]{acl}

\usepackage{times}
\usepackage{latexsym}

\usepackage[T1]{fontenc}

\usepackage[utf8]{inputenc}

\usepackage{microtype}

\usepackage{inconsolata}

\usepackage{graphicx}
\usepackage{colortbl}

\usepackage{booktabs}
\usepackage{amsmath}
\usepackage{amssymb}
\usepackage{physics}
\usepackage{multirow}
\usepackage{enumitem} % Required for noitemsep

\usepackage{tikz}
\usetikzlibrary{positioning}

\usepackage{pgfplots}
\pgfplotsset{compat=1.18}

\title{Beyond statistical significance: Quantifying uncertainty and statistical variability in multilingual and multitask NLP evaluation}

\author{Jonne S{\"a}lev{\"a} \\
  Brandeis University \\
  \texttt{jonnesaleva@brandeis.edu}
  \\\And
  Duygu Ataman \\
  Middle East Technical University \\
  \texttt{dataman@metu.edu.tr}
  \\\And
  Constantine Lignos \\
  Brandeis University \\
  \texttt{lignos@brandeis.edu}
  \\}

\begin{document}
\maketitle
\begin{abstract}
We introduce a set of resampling-based methods for quantifying uncertainty and statistical precision of evaluation metrics in multilingual and/or multitask NLP benchmarks.
We show how experimental variation in performance scores arises from both model and data-related sources, and that accounting for both of them is necessary to avoid substantially underestimating the overall variability over hypothetical replications.
Using multilingual question answering, machine translation, and named entity recognition as example tasks, we also demonstrate how resampling methods are useful for quantifying the replication uncertainty of various quantities used in leaderboards such as model rankings and pairwise differences between models.
\end{abstract}

\section{Introduction}
\label{sec:intro}

Over the last several years, multilingual research has undergone exponential growth within NLP, both in terms of model capabilities as well as evaluation datasets.
Since no paper can evaluate on all languages, it is important to determine to what extent research findings generalize from one language and from one task to another.
This is particularly true in recent years with 
the proliferation of large language models (LLMs), where the goal often is to draw conclusions about the models that are not task or language-dependent.

\begin{figure}
    \centering
    \includegraphics[width=0.9\linewidth]{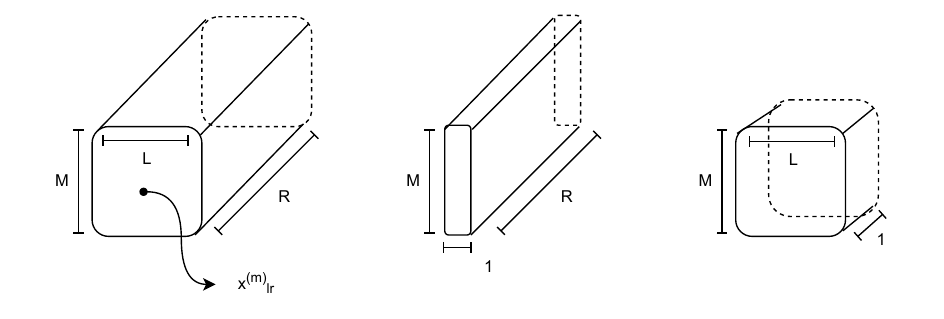}
    \caption{Left: Notional diagram of a $M \times L$ leaderboard consisting of M models tested on L languages and replicated R times each, yielding individual observations $x_{lr}^{(m)}$. Middle: Aggregation over $L$ languages into a single scalar per model and estimating between-language variance $\nu_{m}^2$. Right: Aggregation over replications to estimate within-language uncertainty $\eta_{ml}$.}
    \label{fig:cartoon-plot}
\end{figure}

\paragraph{Evaluation paradigms} Despite this need, evaluation setups in multilingual NLP research tend to focus on within-language evaluation and fall into two broad categories.
In the first category, the same experiments are repeated across several languages and the results analyzed separately.
The inferences are typically reported as one collection, e.g. ``Model A significantly outperforms model B on English--Finnish and English--Turkish but not English--German'' or ``Model A outperforms model B on 13 out of 43 language pairs.''
The second variant, depicted in Figure~\ref{fig:cartoon-plot}, is the ``leaderboard-type'' scenario in which performance is evaluated on each of the $L$ languages after which the numbers are distilled into a single performance measure using a pre-specified aggregation function such as the arithmetic mean.
Given the aggregate scores, the main interest may then be to \textit{rank} the $M$ models based on the single score or to compare differences between observed scores.

\paragraph{Statistical significance and effect sizes}
The most interesting questions to practitioners often revolve around whether a given model ``truly'' outperforms another model or whether an observed performance difference was simply ``random noise'' that could be expected to occur under replication when comparing two equally performing models.
From a statistical perspective, estimating this ``consistency with random noise'' is intimately linked to testing for statistical significance.
Under conventional null hypothesis significance testing the typical ``$p \leq 0.05$'' threshold is achieved when an estimate $\hat{\theta}$ lies at least 1.96 ($\approx 2$) standard errors from a hypothesized null value $\theta_0$, often zero.
Following \citet{gelman2018failure} we refer to the ratio $|\hat{\theta}/\text{se}(\hat{\theta})|$ as \textit{effect size} and use it to judge statistical significance based on whether it exceeds a threshold of 2.

To estimate such effect sizes requires specifying the source(s) of random variation in our experiments.
Typically, this variation has been treated as arising from one of two sources: \textit{data-side variability} and \textit{model-side variability} which we describe below.

\paragraph{Data-side variability}
The former quantifies how differently each model would perform if a slightly different test set were used for a given task.
In practice this is typically done using resampled versions of the original test set, for example using the \textit{bootstrap} resampling algorithm \citep{efron1979bootstrap}.

From a sampling perspective, data-side variability can be interpreted as sampling error associated with the process of constructing a test set for a given task by sampling data from a larger set of all possible test sets.
Another source of data-related variability that appears in leaderboards and other multi-task evaluations is the choice of evaluation tasks. When ranking is performed based on aggregated quantities, the overall performance and rankings of models will vary as a function of which tasks are chosen.
This can be understood as the sampling error associated with creating an evaluation benchmark out of the larger population of all possible ones that could have been constructed.

\paragraph{Model-side variability}
There is also \textit{model-side variability} that will be present even if the data set is held constant.
With LLMs, decoding often involves random sampling of tokens, especially with generative tasks like question answering. By drawing multiple responses for each question, the performance will fluctuate randomly around some average.
Even if the inference algorithm is deterministic (e.g. beam search), any potential finetuning will likely be nondeterministic due to, for example, random shuffling of minibatches and weight initialization. This will yield slightly different parameters from run to run, which will potentially yield different responses and different values of the performance metric.
From a sampling perspective, model-side uncertainty can be seen as drawing samples from the probability distribution over possible responses defined by the model.

\paragraph{Summary}

In this paper, we provide a new perspective on how resampling-based methods can be useful in analysis of experimental results in multilingual and multitask NLP.

Through connections to statistics, we show how experimental variance can be decomposed into between and within-task components ($\nu$ and $\eta_l$) and how the latter further decomposes into model and data-based components ($\sigma_l$, $\tau_l$) which arise naturally from the structure of the experimental data.

Using question answering (QA), machine translation (MT), and named entity recognition (NER) as case studies, our results demonstrate that none of these sources of variation are negligible.
This suggests that only analyzing one source of error may underestimate the total variation and expose researchers to the risk of drawing incorrect conclusions about the relative merits of models.

We show how resampling and estimates of between and within-language variance components can be used to derive uncertainty-aware estimates of more complex quantities such as relative rankings of models and approximate distributions of pairwise performance differences between them.

All of our methods run in seconds to minutes and present no major computational bottlenecks on top of the overall inference time complexity.
We provide an implementation of our approach in a toolkit which we make freely available at \url{https://github.com/j0ma/reuben}.\footnote{The name \texttt{reuben} is an acronym for ``\textbf{RE}sampling-based \textbf{U}ncertainty \textbf{B}ounds for \textbf{E}valuating \textbf{N}LP.}

\section{Related work}

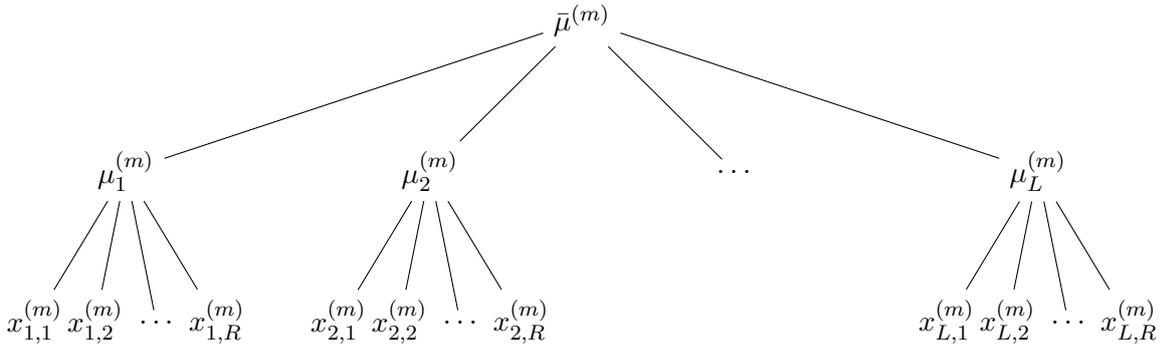
\begin{figure*}[ht]
    \begin{tikzpicture}[
    level 1/.style={sibling distance=40mm},
    level 2/.style={sibling distance=8mm},
    level distance=20mm
]

\node {$\bar{\mu}^{(m)}$}
    child {node {$\mu_1^{(m)}$}
        child {node {$x_{1,1}^{(m)}$}}
        child {node {$x_{1,2}^{(m)}$}}
        child {node {$\ldots$}}
        child {node {$x_{1,R}^{(m)}$}}
    }
    child {node {$\mu_2^{(m)}$}
        child {node {$x_{2,1}^{(m)}$}}
        child {node {$x_{2,2}^{(m)}$}}
        child {node {$\ldots$}}
        child {node {$x_{2,R}^{(m)}$}}
    }
    child {node {$\ldots$}}
    child {node {$\mu_L^{(m)}$}
        child {node {$x_{L,1}^{(m)}$}}
        child {node {$x_{L,2}^{(m)}$}}
        child {node {$\ldots$}}
        child {node {$x_{L,R}^{(m)}$}}
    };
\end{tikzpicture}
    \caption{Tree diagram showing the multilevel structure of the experimental data containing $R$ replications of $L$ languages for a single model.
    Between-language variance $\nu_m^2$ is computed across the averages $\mu^{(m)}_{l}$ whereas
    within-language variance $\eta_{ml}^2$ corresponds to the variance among the leaf nodes $x_{lr}^{(m)}$ of each subtree.
    }
    \label{fig:exp-data-tree}
\end{figure*}

\paragraph{Significance testing in ML and NLP}
Within ML, statistical significance and hypothesis testing have been prevalent since at least the 1990s \citep[e.g.][]{dietterich1998approxtests, dror-etal-2018-hitchhikers,demsar2006statcompmultiple}, though the emphasis has often been on ``finding the right test'' instead of fully modeling the available data.
Within NLP, significance testing is often done using permutation tests and bootstrap-based hypothesis tests \citep{noreen1989computer, koehn-2004-statistical}, which have become the de-facto standard featured in state-of-the-art evaluation toolkits \citep[e.g.][]{post-2018-call}.
Significance testing based on resampling is also often used when evaluating task-specific MT metrics against human judgments in order to gauge what difference magnitudes correspond to real differences as perceived by humans\citep[e.g.][]{lo-etal-2023-beyond,kocmi-etal-2021-ship}. 

In contrast, \citet{ulmer-etal-2022-experimental} advocate for understanding model performance variation based on multiple model training runs instead of resampling, and using this model-side variation to assess statistical significance. There is no clear consensus though, and others \citep[e.g.][]{bethard2022randomseeds} expressly warn against using random seeds to provide estimates of score distributions based on the argument that random seeds should be optimized like other hyperparameters. 
Some work in NLP has also evaluated different sources of variation in evaluation of MT metrics \citep{xiang-etal-2022-investigating}.

\paragraph{Evaluation practices in NLP}

There has also been a healthy debate regarding best practices in test set construction and more generally of evaluation using held-out datasets. 
\citet{gorman-bedrick-2019-need} and others \citep[e.g.][]{kodner-etal-2023-morphological, liu-dorr-2024-effect} argue that we should \textit{randomize} our splits and evaluate on several versions of our test sets to reduce the chance of false positives.
    \citet{sogaard-etal-2021-need} instead favor using approaches based on adversarial splitting, arguing that naive randomization will underestimate the variation.

Benchmarks have received much attention in the general machine learning community. An excellent survey is \citet{benchmarklottery2021} where the authors discuss the lifecycle of overall ML benchmarks, how they become stale as well as what is random when models are evaluated.
The question of what aggregation function to use in leaderboards is also explored by \citet{tatiana2021not} where the authors explore using both the arithmetic, geometric and harmonic means. In particular, they show that results may be wildly different depending on what aggregation measure is used.
\citet{longjohngopalan2025statisticaluq} also survey and develop resampling-based and Bayesian methods for popular computer vision benchmarks for LLMs.

\paragraph{Related ideas in statistics}

While the original bootstrap algorithm \citep{efron1979bootstrap} was intended as a frequentist estimation device, a Bayesian interpretation was provided by \citet{rubin1981bayesboot}, allowing the resulting estimates (empirical distributions, intervals etc.) to be interpreted as posterior quantities.
Uncertainty quantification (i.e. standard error estimation) using bootstrap and other nonparametric methods is thoroughly reviewed by \citet{efron1981nonparametric}.
On the importance of understanding variation, \citet{gelman2005anova} provides an excellent argument for the importance of understanding variance components in statistical analysis more generally. 
For a more specific application to linguistic research, see \citet{vasishth2021embrace}.
The multilevel structure we use is also related to the technique of \textit{random-effects meta-analysis}, see \citet{rema2008}.

\paragraph{Uncertainty quantification in NLP}

There exists a substantial literature \citep[e.g.][]{nikitin2024kernellanguageentropy, da2025understanding, chen2024questionrephrasing, chen2025uq, yang-etal-2025-maqa, ye2024benchmarking, wagner2024blackbox, blackwell2025towardsrepro} on a task referred to as \textit{uncertainty quantification} in NLP, where the task is to estimate some properties of an LLM's distribution over continuations, $p(x|x_{\text{prompt}})$, that predicts whether the model is able to produce the correct answer to this question. While this task quantifies per-example uncertainty, our efforts instead focus on \textit{experimental variation}. i.e. estimating the noise we would expect to see under repeated experiments and hypothetical evaluation suites. 

\section{Statistical background}

We begin by formalizing the analysis of multilingual experimental results as a statistical inference problem related to a hierarchically structured population of experimental results.

\paragraph{Observations}

Our experimental data consists of scores generated by separately evaluating $M$ models on $L$ languages. Individual model-level observations are represented as $L$-vectors of scores:
$$\vb{x}^{(m)} = (x^{(m)}_1, \dots, x^{(m)}_L)^\top \in \mathbb{R}^L$$
The entire table of experimental results is then an $M$-by-$L$ matrix where each row corresponds to a model $\vb{x}^{(m)}$, as shown in Figure~\ref{fig:cartoon-plot}.
Hierarchically, our data can be seen as arising from a three-level population 
$\bar{\mu}^{(m)} \mapsto \mu^{(m)}_l \mapsto x^{(m)}_{lr}$.
This structure is depicted in Figure~\ref{fig:exp-data-tree} in the Appendix.
At the topmost level, each model is associated with a population-level average performance ($\bar{\mu}^{(m)}$), and individual languages are seen as subsets of this larger population, each with their own averages ($\mu^{(m)}_l$). At the lowest level, we have the observed scores ($x^{(m)}_{lr}$) which represent replications of an experiment on a given subpopulation.
Leaderboard-style evaluation with aggregate performance measures can then be seen as estimating $\bar{\mu}^{(m)}$ by using an aggregation function as the estimator.

\paragraph{Sources of uncertainty}
Under infinite replications, our observed values $x^{(m)}_l$ values will center around some mean value and exhibit a degree of fluctuation around it, representing measurement error arising from both model-side randomness (e.g. nondeterministic decoding) as well as sampling error of the test set.
We can represent this as $x^{(m)}_{lr} = \mu^{(m)}_l + \varepsilon^{\text{repl}}_{lr}$
where $\varepsilon^{\text{repl}}_{lr}$ represents the deviation of observed scores from the language-specific mean $\mu^{(m)}_l$.
In addition to this ``within-language'' variation arising from replication, there is also ``between-language'' variability related to what tasks are included in the leaderboard. Concretely, each $\mu^{(m)}_l$ also differs from the global mean performance across all tasks by some amount, i.e. $\mu^{(m)}_l = \bar{\mu}^{(m)} + \varepsilon^{\text{lang}}_l$. 
This lets us decompose the overall observations into
$x^{(m)}_{lr} = \bar{\mu}^{(m)} + \varepsilon^{\text{lang}}_l + \varepsilon^{\text{repl}}_{lr}$.
Taking the variance of both sides, we obtain a first-principles variance decomposition into between and within-language components
$
\mathbb{V}[x^{(m)}_{lr}] = \mathbb{V}[\varepsilon^{\text{lang}}_l] +\mathbb{V}[\varepsilon^{\text{repl}}_{lr}] = \nu_m^2 + \eta^2_{lm}
$
where $\nu_m^2$ and $\eta^2_{lm}$ refer to between and within-language variation of model $m$, respectively.

\paragraph{Replication}
For a given task, two orthogonal sources of replication noise are apparent. \textit{Model-side randomness} that arises from nondeterminism in decoding/text generation.
Additional model-side randomness may arise from random parts of any training/finetuning runs, related to randomized weight initialization and batching.
While less common in inference-only LLM evaluation, this is still a relevant source of variability for older non-pretrained models such as LSTMs \citep[e.g.][]{reimers-gurevych-2017-reporting} and older pretrained models such as BERT and XLM-R.
Another common source of score variability is \textit{sampling error} when constructing each test set $\mathcal{D}_{l}$ \citep[e.g.][]{koehn-2004-statistical, benchmarklottery2021}.
Since we tend not to have access to true sampling distribution $p(\mathcal{D}_l)$, this variability is typically estimated by \textit{resampling} with replacement from the original data and computing an empirical variance estimate.
This is the approach taken by many evaluation libraries such as \texttt{sacrebleu} \citep{post-2018-call} and \texttt{lm-evaluation-harness} \citep{eval-harness}.

From a sampling perspective, these sources of variability can be viewed as parts of the data collection process: first, given a model $m$ and a random seed $s$, we sample random responses to all examples in our test set, i.e. $x_{ls} \sim p(x_{l})$. 
Finetuning can be seen as sampling $\theta^{(m)}_{ls} \sim p(\theta^{(m)}_l)$ from the distribution of all model parameters that could be obtained.
On the dataset side, we sample a set of languages to evaluate on and, for each language, a test set from a larger hypothetical population of similar data in language $l$, $\mathcal{D}_{lb} \sim p(\mathcal{D}_l)$.
This decomposes the total variance further as
$$
\mathbb{V}[x^{(m)}_{lr}]
= \underbrace{\nu_m^2}_{\text{lang}} + \underbrace{\sigma_{ml}^2}_{\text{seed}} + \underbrace{\tau_{ml}^2}_{\text{boot}}
$$
In our analysis, we use both seed and bootstrap-based resampling to yield a total of $R = SB$ replications per language.
While the $B$ resamples may be much cheaper to obtain than the $S$ model re-instantiations in terms of time cost, we feel that using both is helpful to properly estimate the decomposition as well as to understand whether the sources contribute equally to the total within-language variance for each language.

\paragraph{Why care about variance components?}

Most immediately, estimates of $\eta_l$ (total within-language variability), $\sigma_l$ (model-side variability), and $\tau_l$ (test data variability) provide the researcher with estimates of what is driving the variation in their data. 
High values of $\sigma_l^2$ (model-side variability) suggest that the learned distribution over responses may have high entropy or, in the case of finetuning, that the architecture may be highly sensitive to random shuffling of batches due to e.g. small amounts of training data. 
On the other hand, high values of $\tau_l^2$ (test data variability) and $\nu_m^2$ (between-language variance) indicate that a model's performance is particularly sensitive to the exact composition of a test set or benchmark and may suggest lower ability to generalize due to overfitting.
Variance components are also tied to statistical significance: we can, roughly speaking, judge an estimate as being statistically significant if it lies more than two standard errors from zero \citep[e.g.][]{gelman2018failure}.

\paragraph{Estimating model and data-side SD}

Using the predictions of $S$ replications on the original test data (i.e. no bootstrapping), we estimate $\sigma_l$ using the sample standard deviation formula.
Since we have observations of each of the seeds on all $B$ datasets, we also compute an estimate of the standard error $\hat{\text{se}}(\hat{\sigma}_l)$ using the sample standard deviation formula computed over the bootstrap datasets.

To estimate the boot-to-boot variability $\tau_l$, we first compute the F1 score variance over the $B$ bootstrap datasets separately for each of the $S$ seeds.

We then construct the average estimator $
\bar{\tau}_l = \sum_{s=1}^S \hat{\tau}_{ls}/S
$.
The standard error of $\hat{\tau}_l$ is estimated using the usual sample standard deviation formula over seeds.
This gives us the plug-in estimate of the standard error of the averaged estimator  
$\hat{\text{se}}(\bar{\tau}_l) = \hat{\text{se}}(\hat{\tau}_{ls})/\sqrt{S}$.

\paragraph{Aggregate scores and their standard errors}
In most leaderboard-style scenarios, aggregated scalar performance measures such as the arithmetic mean $\bar{x}^{(m)}_{1:L} = \sum_l x^{(m)}_l/L$ tend to be used instead of the full score vectors.
Its popularity of the arithmetic mean can be explained by how simple it is to compute, as well as the closed-form expression for its standard deviation $\text{sd}(x^{(m)}_1, \dots, x^{(m)_L})/\sqrt{L}$ using only the between-language SD $\nu_m$ and the within-language SDs $\eta_{ml}$.
Other aggregation functions, such as the geometric mean, or median may also be used, although they may not be as well-behaved in terms of standard error.
For such aggregation functions, the SD typically does not exist in closed form and must be estimated using resampling.

\section{Task 1: Question answering with LLMs}

As our first case study, we focus on multilingual question answering and evaluate four LLMs on all subsets of XQuAD \citep{artetxe-etal-2020-cross}. 
Specifically, we use \texttt{aya-expanse-8B} \citep{dang2024ayaexpansecombiningresearch}, \texttt{TowerInstruct-Mistral-7B-v0.2} \citep{alves2024tower}, Google's \texttt{gemma2-9b} and finally \texttt{Clarus-7B-v0.3}. 
We evaluate using token-level F1 score, using the standard \texttt{lm-evaluation-harness} implementation. 
Details of the corpus, experimental settings and hardware used are in Section~\ref{xquad-appendix} in the Appendix.

\subsection{Variance components}
The \texttt{lm-evaluation-harness} library we use to run our  experiments automatically computes bootstrapped standard errors for F1.
In addition to the bootstrap SE, we incorporate model-side uncertainty by sampling 5 answers for each question from each model.
The summarized variance components are displayed in Table~\ref{tab:varcomp-xquad}. The full decomposition is available in Table~\ref{tab:varcomp_table_xquad_full} in the Appendix.

\begin{table}[htb]
    \centering
    \footnotesize
    \begin{tabular}{lrrrr}
\toprule
Model & Mean & SD & Min & Max \\
\midrule
\multicolumn{4}{l}{\textit{Between-language} ($\nu$)} & \\
\midrule
Clarus 7B & 8.92 & - & - & - \\
TowerInstruct 7B & 11.03 & - & - & -\\
Aya Expanse 8B & 12.01 & - & - & - \\
Gemma 2 9B & 5.05 & - & - & - \\
\midrule
\multicolumn{4}{l}{\textit{Total within-language} ($\eta_{ml}$)} & \\
\midrule
Clarus 7B & 0.89 & 0.41 & 0.40 & 1.67 \\
TowerInstruct 7B & 0.73 & 0.32 & 0.19 & 1.21 \\
Aya Expanse 8B & 1.47 & 0.18 & 1.18 & 1.73 \\
Gemma 2 9B & 1.21 & 0.16 & 0.93 & 1.51 \\
\midrule
\multicolumn{4}{l}{\textit{Boot-to-boot} ($\tau_{ml}$)} & \\
\midrule
Clarus 7B & 0.70 & 0.29 & 0.24 & 1.27 \\
TowerInstruct 7B & 0.59 & 0.28 & 0.15 & 1.05 \\
Aya Expanse 8B & 1.24 & 0.11 & 1.06 & 1.42 \\
Gemma 2 9B & 0.95 & 0.10 & 0.80 & 1.18 \\
\midrule
\multicolumn{4}{l}{\textit{Seed-to-seed} ($\sigma_{ml}$)} & \\
\midrule
Clarus 7B & 0.53 & 0.33 & 0.16 & 1.25 \\
TowerInstruct 7B & 0.41 & 0.18 & 0.12 & 0.74 \\
Aya Expanse 8B & 0.76 & 0.30 & 0.25 & 1.23 \\
Gemma 2 9B & 0.74 & 0.20 & 0.48 & 1.09 \\
\bottomrule
\end{tabular}
    \caption{Summary of variance components for QA experiments.}
    \label{tab:varcomp-xquad}
\end{table}

\begin{table*}[htb]
    \centering
    \footnotesize
    \begin{tabular}{lrrrrrr}
\toprule
 & \multicolumn{3}{c}{Clarus 7B} & \multicolumn{2}{c}{TowerInstruct 7B} & Aya Exp. 8B \\
 \cmidrule(lr){2-4} \cmidrule(lr){5-6} \cmidrule(lr){7-7}
 & TowerInstruct 7B & Aya Exp. 8B & Gemma 2 9B & Aya Exp. 8B & Gemma 2 9B & Gemma 2 9B \\
\midrule
Arabic & 8.75 $ \pm $ 1.13 & -24.27 $ \pm $ 1.34 & -0.11 $ \pm $ 1.50\textbf{*} & -33.02 $ \pm $ 1.39 & -8.86 $ \pm $ 1.53 & 24.16 $ \pm $ 1.71 \\
Chinese & -3.96 $ \pm $ 1.50 & -22.80 $ \pm $ 1.66 & -6.34 $ \pm $ 1.47 & -18.83 $ \pm $ 1.84 & -2.38 $ \pm $ 1.67\textbf{*} & 16.45 $ \pm $ 1.82 \\
English & 5.21 $ \pm $ 1.67 & -33.30 $ \pm $ 1.86 & 8.79 $ \pm $ 1.95 & -38.50 $ \pm $ 1.67 & 3.58 $ \pm $ 1.75 & 42.09 $ \pm $ 1.95 \\
German & -6.43 $ \pm $ 1.41 & -32.16 $ \pm $ 1.44 & -9.04 $ \pm $ 1.46 & -25.73 $ \pm $ 1.72 & -2.61 $ \pm $ 1.74\textbf{*} & 23.12 $ \pm $ 1.77 \\
Greek & 16.45 $ \pm $ 1.12 & -20.77 $ \pm $ 1.69 & 1.02 $ \pm $ 1.58\textbf{*} & -37.22 $ \pm $ 1.59 & -15.43 $ \pm $ 1.47 & 21.79 $ \pm $ 1.94 \\
Hindi & 20.43 $ \pm $ 0.94 & -26.75 $ \pm $ 1.51 & 1.26 $ \pm $ 1.46\textbf{*} & -47.18 $ \pm $ 1.34 & -19.17 $ \pm $ 1.27 & 28.01 $ \pm $ 1.76 \\
Romanian & 2.36 $ \pm $ 1.45\textbf{*} & -30.14 $ \pm $ 1.55 & -3.32 $ \pm $ 1.81\textbf{*} & -32.50 $ \pm $ 1.69 & -5.69 $ \pm $ 1.95 & 26.81 $ \pm $ 2.01 \\
Russian & -6.20 $ \pm $ 1.59 & -20.23 $ \pm $ 1.65 & 5.36 $ \pm $ 1.91 & -14.02 $ \pm $ 1.65 & 11.57 $ \pm $ 1.92 & 25.59 $ \pm $ 1.96 \\
Spanish & 8.01 $ \pm $ 2.05 & -17.36 $ \pm $ 1.98 & 12.55 $ \pm $ 2.23 & -25.37 $ \pm $ 1.70 & 4.54 $ \pm $ 1.99 & 29.91 $ \pm $ 1.92 \\
Thai & 21.01 $ \pm $ 1.23 & -3.83 $ \pm $ 1.83 & 6.42 $ \pm $ 1.69 & -24.84 $ \pm $ 1.52 & -14.60 $ \pm $ 1.35 & 10.25 $ \pm $ 1.90 \\
Turkish & 7.06 $ \pm $ 0.90 & -25.13 $ \pm $ 1.50 & -7.01 $ \pm $ 1.38 & -32.19 $ \pm $ 1.53 & -14.07 $ \pm $ 1.40 & 18.12 $ \pm $ 1.85 \\
Vietnamese & 3.23 $ \pm $ 1.06 & -29.35 $ \pm $ 1.41 & -5.82 $ \pm $ 1.44 & -32.59 $ \pm $ 1.39 & -9.05 $ \pm $ 1.41 & 23.53 $ \pm $ 1.68 \\
\bottomrule
\end{tabular}

    \caption{Pairwise differences between models on XQuAD. Non-significant differences are indicated with an asterisk.}
    \label{tab:pairwise-diffs-xquad}
\end{table*}

\subsection{Resampling for model comparison}
We seek to quantify the uncertainty in the XQuAD F1 scores in each language as well as aggregated across languages.
Specifically, we compute pairwise model differences on each language as well as between aggregate scores computed using the arithmetic mean, geometric mean and median. We also compute rankings as point estimates.

We estimate the difference between models using the following parametric bootstrap resampling procedure.
Given a language, we first take an estimate of each model's average performance on it by averaging the $S = 5$ observed values. We then use the estimate of the within-language SD, $\eta_{ml}$, to get a randomly resampled performance score for each model.

We estimate $\eta_{ml}$ by first computing the standard deviation of scores across seeds and then use the average bootstrap SD $\tau_{ml}$ across the $S=5$ values computed by \texttt{lm-evaluation-harness} to form an aggregate SD estimate of $\eta_{ml}$. We then sample a noise term using a random draw from $\varepsilon \sim \mathcal{N}(0, \eta_{ml}^2)$ yielding a resampled score $x + \varepsilon$ with variance $\eta_{ml}^2 = \sigma_{ml}^2 + \tau_{ml}^2$. 
While the score itself may not be Gaussian, we believe that modeling the noise terms as such is reasonable, as we observe mostly small score deviations across both seeds and bootstrap resamples.

We repeat this process $R=1000$ times. For each replication, we compute a performance difference on each language, and an aggregated score of performance scores across languages, using the arithmetic mean, geometric mean and median. We also rank the models based on the randomly replicated performance scores. 
The results of this procedure are summarized in several tables below. 

Table~\ref{tab:pairwise-diffs-xquad} summarizes pairwise differences between models and their associated SDs. Table~\ref{tab:xquad-ranks-by-summary-stat} gives the rank distribution of each model using each of the summary statistics.
While Aya Expanse 8B and TowerInstruct 7B reliably place first and last regardless of metric, the rank distributions of Clarus 7B and Gemma 2 9B can vary significantly depending on which aggregation function is used.

\begin{table}[htb]
    \centering
    \footnotesize
    \begin{tabular}{lrrrr}
\toprule
& \multicolumn{4}{c}{Arithmetic mean} \\
\midrule
& Aya 8B & Clarus 7B & Gemma 2 9B & TI 7B \\
\midrule
1 & \textbf{100.00} & - & - & - \\
2 & - & 24.50 & \textbf{75.50} & - \\
3 & - & \textbf{75.50} & 24.50 & - \\
4 & - & - & - & \textbf{100.00} \\
\midrule
& \multicolumn{4}{c}{Median} \\
\midrule
1 & \textbf{100.00} & - & - & - \\
2 & - & 35.30 & \textbf{64.70} & - \\
3 & - & \textbf{64.70} & 35.30 & - \\
4 & - & - & - & \textbf{100.00} \\
\midrule
& \multicolumn{4}{c}{Geometric mean} \\
\midrule
1 & \textbf{100.00} & - & - & - \\
2 & - & \textbf{94.90} & 5.10 & - \\
3 & - & 5.10 & \textbf{94.90} & - \\
4 & - & - & - & \textbf{100.00} \\
\bottomrule
\end{tabular}

    \caption{Distribution of model ranks on XQuAD using different aggregators. Simulation computed over 1,000 randomly sampled performance scores.}
    \label{tab:xquad-ranks-by-summary-stat}
\end{table}

\section{Task 2: Machine translation with LLMs}

As a second case study, we seek to compare the performance of three translation-oriented LLMs on the \texttt{devtest} split of the multilingual FLORES-200 translation benchmark \citep{nllb-24}.
Specifically, we use \texttt{aya-expanse-8B} \citep{dang2024ayaexpansecombiningresearch}, \texttt{TowerInstruct-Mistral-7B-v0.2} \citep{alves2024tower} (both derived from Mistral 7B), and Clarus 7B.
Our reasoning for choosing these models despite the small parameter count of $\sim$7B is twofold. First, these models are some of the few that are either evaluated on or specifically engineered for translation. Second, our computational resources limit the size of models we can run without quantization.
Ideally, we would also test on all languages covered by FLORES-200, but the language support of the LLMs restricts us to Dutch, English, French, German, Italian, Korean, Portuguese, Russian, and Spanish.
For each language we create two language pairs: XX$\rightarrow$EN and EN$\rightarrow$XX.
We evaluate using BLEU \citep{papineni-etal-2002-bleu},  ChrF++ \citep{popovic-2015-chrf} and COMET \citep{rei-etal-2020-comet}.

\subsection{Variance components}

The \texttt{lm-evaluation-harness} library we use to run our  experiments automatically computes bootstrapped standard errors for BLEU, ChrF++ and COMET.
In addition to the bootstrap SE, we incorporate model-side uncertainty by sampling 5 translation hypotheses for each source sentence from each model.
The summarized results are shown in Table~\ref{tab:flores_varcomp_table} in the Appendix. 
A detailed variance decomposition of the within-language variation is shown in Table~\ref{tab:varcomp_table_flores_full} in the Appendix.
As with XQuAD, the between-language variability ($\nu$) is much larger than either the model-side or bootstrap uncertainty.
Most variance components are also less than 1.0 BLEU or ChrF++ points.

\subsection{Resampling for model comparison}

As with XQuAD, we also wish to quantify the uncertainty in the performance differences between the two models on each language.
Given observed BLEU/ChrF++/COMET scores for each language, we can easily compute a point estimate by subtraction.
We quantify the uncertainty in this point estimate by resampling as with XQuAD.
The estimated differences are displayed in Table~\ref{tab:flores_results_uncert_seedboot}.
The most obvious observation is that Clarus 7B clearly underperforms relative to Aya Expanse 8B and TowerInstruct 7B. 
The latter two models perform more similarly to each other and significant differences were found in 6 out of 16 translation using BLEU. 
Out of the 6 significant differences, TowerInstruct 7B beat Aya Expanse 8B 4 out of 6 times. 
When using ChrF++, significant differences were found in 10 out of 16 translation tasks, with TowerInstruct 7B leading all of them.

\section{Task 3: NER}\label{varcomp-casestudy}

As a third task, we show how resampling can be used for model comparison in multilingual NER when finetuning pretrained language models.
Our analysis is based on the OpenNER 1.0 multilingual NER benchmark \citep{palen-michel-etal-2025-openner} which consists of 61 unique datasets covering a total of 51 languages. Since we reanalyze the results of the original paper, we use the same models: mBERT and XLM-R.
For each model architecture, we vary the training data according to two conditions on each language-dataset pair: (i) \texttt{individual}, where each model is finetuned with data from only a given training set and (ii) \texttt{concatenated}, where each model is finetuned with a concatenated and downsampled version of all the training data included in OpenNER.
In addition to the original results, we construct $B=100$ bootstrap replicated data sets and reuse the same predictions from the original test set. 
This gives us a total of $R = SB = 1000$ replicated scores for each of the 61 datasets.

\subsection{Variance components via resampling}\label{sec:varcomp-ner}

Like with QA and MT, we investigate how resampling-based inference can be used to better make sense of experimental variation by estimating the variance components due to model and data-side uncertainty for each of the 61 languages pairs that comprise the overall benchmark.
A concise display of inferences for $\nu$, $\sigma$ and $\tau$  are shown in Table~\ref{tab:ner-varcomp-summary-smalltable-maintext}. 
A more detailed table of estimates is shown in the Appendix in Table~\ref{tab:varcomp_table}.

\begin{table}[tb]
    \centering
    \small
    % All non-math mode
\begin{tabular}{llrr}
\toprule
& & \multicolumn{2}{c}{Finetuning data} \\
\cmidrule(lr){3-4}
& Model & Individual & Concatenated \\
\midrule
& Glot500 & 8.27 & 6.30 \\
 $\nu$ & mBERT   & 9.95 & 14.38 \\
& XLM-R   & 7.13 & 6.31 \\
\midrule
& Glot500 & 1.59\textsubscript{(2.51)} & 0.61\textsubscript{(0.39)} \\
$\sigma_l$ & mBERT   & 1.15\textsubscript{(2.55)} & 0.81\textsubscript{(0.45)} \\
& XLM-R   & 1.24\textsubscript{(3.53)} & 0.64\textsubscript{(0.42)} \\
\midrule
& Glot500 & 1.27\textsubscript{(0.95)} & 1.17\textsubscript{(0.79)} \\
$\tau_l$ & mBERT   & 1.36\textsubscript{(0.86)} & 1.47\textsubscript{(0.91)} \\
& XLM-R   & 1.23\textsubscript{(0.84)} & 1.18\textsubscript{(0.79)} \\
\bottomrule
\end{tabular}

    \caption{Summary of variance components for NER. The three groups of rows, $\nu$, $\sigma$ and $\tau$ correspond to between-language, seed, and bootstrap SDs. Subscripts indicate standard deviations across languages.}
    \label{tab:ner-varcomp-summary-smalltable-maintext}
\end{table}

Looking at Table~\ref{tab:ner-varcomp-summary-smalltable-maintext}, we see that both model and data-side variability tend to lie in the 0.6--1.6 F1 range. 
When we compare models in the individual versus the concatenated conditions, the models in the concatenated condition exhibit distinctly lower model-side variability.
This suggests that languages with less training data benefit from larger, multilingual finetuning, even if no positive transfer learning is taking place, possibly due to a reduction in gradient instability.

Model-side variability, on the other hand, remains relatively constant across individual and concatenated conditions which suggests that the size of finetuning sets may not help much in terms of dataset-to-dataset generalization.
This observation also highlights how the model and data-side variance components measure fundamentally different aspects of performance variability and how both are needed for thorough performance assessment. 

\subsection{Resampling for model comparison}

As with QA and NMT, we also estimate quantities other than variance components. To estimate uncertainties for model rankings and pairwise differences between models, we resample the $R=SB=1000$ replications within each language.
Unlike our QA and MT experiments, we do this entirely nonparametrically by sampling a seed and replication index at random.
We also show how the between-language score variability ($\nu$) can be incorporated into the analysis by also resampling what languages are considered when estimating overall performance.

\paragraph{Pairwise differences}

\begin{table*}[t]
    \centering
    \small
    % 3x1 layout
\begin{tabular}{lrrrrr}
\toprule
\multicolumn{6}{c}{Mean: $\bar{\Delta}$} \\
\midrule
$\textit{Model}$ & Glot500 (i) & mBERT (c) & mBERT (i) & XLM-R (c) & XLM-R (i) \\
\midrule
Glot500 (c) & 2.01 & 12.24 & 8.63 & 0.79 & 2.00 \\
Glot500 (i) &  & 10.23 & 6.63 & -1.22 & -0.01 \\
mBERT (c) &  &  & -3.60 & -11.45 & -10.24 \\
mBERT (i) &  &  &  & -7.84 & -6.63 \\
XLM-R (c) &  &  &  &  & 1.21 \\
\toprule
\multicolumn{6}{c}{Effect size: $\bar{\Delta}/\text{se}(\bar{\Delta})$} \\
\midrule
\textit{Original} & Glot500 (i) & mBERT (c) & mBERT (i) & XLM-R (c) & XLM-R (i) \\
\midrule
Glot500 (c) & 3.97 & 41.70 & 18.73 & 2.43 & 3.07 \\
 Glot500 (i) &  & 23.57 & 10.54 & -2.38 & -0.17 \\
 mBERT (c) &  &  & -9.99 & -38.85 & -19.21 \\
 mBERT (i) &  &  &  & -16.49 & -9.61 \\
 XLM-R (c) &  &  &  &  & 1.78 \\
\midrule
\textit{Subsampled} & Glot500 (i) & mBERT (c) & mBERT (i) & XLM-R (c) & XLM-R (i) \\
\midrule
Glot500 (c) & 1.35 & 3.61 & 4.14 & 1.15 & 1.22 \\
Glot500 (i) &  & 2.52 & 3.25 & -0.82 & -0.00 \\
mBERT (c) &  &  & -0.73 & -3.22 & -2.42 \\
mBERT (i) &  &  &  & -4.05 & -3.31 \\
XLM-R (c) &  &  &  &  & 0.81 \\
\bottomrule
\end{tabular}
    \caption{Means and effect sizes of pairwise F1 score differences $\Delta_{m} = \hat{\mu}^{(m)}_{\text{arith}} - \hat{\mu}^{(n)}_{\text{arith}}$. 
    }
    \label{tab:pairwise-diffs-effect-size}
\end{table*}

To obtain estimates of the pairwise differences of average performance between models, $\Delta_{mn} = \hat{\mu}^{(m)}_{\text{arith}} - \hat{\mu}^{(n)}_{\text{arith}}$, we compute a pairwise difference matrix for each replication $r$.
Given the $M\times M\times R$-dimensional array of pairwise differences, we average over the $R$-dimension to obtain a single $M \times M$ table of average differences.
This is displayed as the top table in Table~\ref{tab:pairwise-diffs-effect-size}. 

To account for uncertainty, we also compute a standard deviation of each pairwise difference across the $R$-dimension.
We then divide the estimated mean difference by the SD to obtain an estimate of the effect size which allows us to assess the statistical significance of the observed differences.
These effect sizes are displayed in the second part of Table~\ref{tab:pairwise-diffs-effect-size}. 
Except for the differences between XLM-R (ind.) and Glot500 (ind.), as well as XLM-R (conc.) and XLM-R (ind.), all estimates appear to be significantly different from zero.

We also conduct another simulation where we subsample $10$ datasets without replacement and otherwise estimate effect sizes as above.
The motivation for this is to incorporate between-language variability into the estimation of overall performance.
Intuitively, this also corresponds to a simulating the effect of downstream usage of the tested models on smaller sets of tasks.
These results are shown in the bottom part of Table~\ref{tab:pairwise-diffs-effect-size}.
Overall, the effect size estimates are much smaller, with many shrinking below magnitude 2 due to the increased standard error.
This suggests that using fixed benchmarks may understate the differences between two models .
While the chosen threshold for significance is arbitrary, the pattern of effect size shrinkage due to the increased variability is robust.

\begin{table}[ht]
    \centering
    \footnotesize
    \begin{tabular}{lrrrrrr}
\toprule
& \multicolumn{2}{c}{Glot500} & \multicolumn{2}{c}{mBERT} & \multicolumn{2}{c}{XLM-R} \\
\cmidrule(lr){2-3} \cmidrule(lr){4-5} \cmidrule(lr){6-7}
& Conc. &  Ind. & Conc. &  Ind. & Conc. & Ind. \\
\midrule
1 & \textbf{99\%} & - & - & - & 1\% & - \\
2 & 1\% & 1\% & - & - & \textbf{97\%} & 1\% \\
3 & - & 44\% & - & - & 2\% & \textbf{54\%} \\
4 & - & \textbf{55\%} & - & - & - & 45\% \\
5 & - & - & 1\% & \textbf{99\%} & - & - \\
6 & - & - & \textbf{99\%} & 1\% & - & - \\
\bottomrule
\end{tabular}

    \caption{Bootstrap-resampled frequency distribution of ranks for each model using arithmetic mean as an aggregation function.}
    \label{tab:ranktable}
\end{table}

\paragraph{Ranks}
To estimate a distribution for model rankings, we recompute the model ranks using the arithmetic mean across languages for each replication.
This yields an empirical distribution estimate of model ranks which is displayed in Table~\ref{tab:ranktable}.
The bolded elements indicate the ranks observed using the original data.
Overall, it seems like the ranks are stable for most models, except for Glot500 (ind.) and XLM-R (ind.) which compete for the 3rd place.
We also repeated the simulation with fixed languages but observed that the ranks remained constant. 
This is in line with the observation from Section~\ref{sec:varcomp-ner} that the between-language variation comprises by far the largest variance component.
We suspect that under the ``subsampling'' condition, the ranks would display a lot more variability but leave this for future work.

\section{Conclusion}

In this paper, we have shown how replication and resampling can be helpful tools in the evaluation of multilingual NLP models.
Through experiments on question answering, machine translation, and named entity recognition, we showed how resampling-based inference can be used to estimate different \textit{variance components} which describe how robust model performance may be to model-related randomness as well as test set composition and between-language variation.
We also showed that the between-language component may dominate the standard error of estimators such as the arithmetic mean in leaderboard-style evaluation settings that involve multiple languages/tasks.

In addition to variance components, we also showed how resampling can be used to estimate the distributions and standard errors of quantities with no closed form expressions, such as pairwise differences between models and rank distributions.
We also explained how standard errors relate to statistical significance and showed that underestimated SEs may lead to overly optimistic 
results that ultimately fail to replicate.

We also wish to stress stress that the techniques introduced in this paper need not translate to significant additional computational overhead.
In particular, between-language/task variance can be computed without any resampling and within-test set bootstrap resampling procedures can be run in seconds on a modern consumer CPU.
While model-side variance may require comparatively more resources in some cases, training on multiple seeds is often only relevant when working with  smaller, non-LLM models. 
With LLMs, it is sufficient to sample multiple responses for a given input which can be done at the fraction of the cost of an entire retraining run.

It is our hope that this paper can stimulate further research into understanding the sources, extent, and nature of performance variation in empirical NLP research.
We conclude with a vision for uncertainty-aware multilingual/multitask evaluation in NLP research. Researchers should:

\begin{enumerate}[wide, labelwidth=!, labelindent=0pt]

\item Thoughtfully consider all relevant sources of randomness when evaluating and comparing different models against each other. Draw more than one set of predictions to understand how variable the performance of a model is, and use bootstrapping to complement model-side uncertainty with data-side uncertainty estimates.

\item In multilingual/multitask evaluation setups, strive to understand how observed differences between models vary from task to task. Incorporate estimates of between-language variation into statistical comparisons to avoid false positives that may not be replicable.

\item If model performance or observed differences are very noisy, communicate this clearly and describe the differences between tasks. This is preferable to attempting to distill results to a binary judgment about which model is better. If feasible, show example cases where models agree/disagree.
    
\item Use resampling with the replications obtained from the above steps to estimate distributions and associated uncertainty for downstream quantities of interest, such as pairwise differences and model rankings.

\end{enumerate}

\section*{Limitations}

While the total number of languages we investigate is quite large, the languages are not evenly distributed across the tasks we study.
Due to the length limit of this venue and the limitations of existing work, we are only able to present analysis of three tasks in this paper.
We have chosen tasks with 100\% human-generated annotation because we believe the results could be skewed if we include datasets generated via system output \citep[e.g.][]{pan-etal-2017-cross}, including LLM-generated annotation or automatically translated datasets.

\section*{Ethical considerations}

As a position paper, this paper argues for more careful evaluation of NLP experiments, particularly when multiple languages and models are studied in parallel.
We believe this can be beneficial to the larger NLP community as it may help practitioners avoid drawing poorly supported conclusions and thus avoid making non-generalizable claims about research findings.
That said, our methods are slightly more involved than traditional methods (e.g. hypothesis testing) and may theoretically lead to more conservative statistical analysis which may in theory obscure real effects.

\section*{Acknowledgments}
This work was supported by the grant \emph{Improving Relevance and Recovery by Extracting Latent Query Structure} by eBay to Brandeis University.
This work was also supported by Brandeis University through internal research funds.

\bibliography{custom,anthology-1,anthology-2}

\begin{thebibliography}{49}
\providecommand{\natexlab}[1]{#1}

\bibitem[{Alves et~al.(2024)Alves, Pombal, Guerreiro, Martins, Alves, Farajian, Peters, Rei, Fernandes, Agrawal, Colombo, de~Souza, and Martins}]{alves2024tower}
Duarte~Miguel Alves, Jos{\'e} Pombal, Nuno~M Guerreiro, Pedro~Henrique Martins, Jo{\~a}o Alves, Amin Farajian, Ben Peters, Ricardo Rei, Patrick Fernandes, Sweta Agrawal, Pierre Colombo, Jos{\'e} G.~C. de~Souza, and Andre Martins. 2024.
\newblock \href {https://openreview.net/forum?id=EHPns3hVkj} {{Tower: An Open Multilingual Large Language Model for Translation-Related Tasks}}.
\newblock In \emph{First Conference on Language Modeling}.

\bibitem[{Artetxe et~al.(2020)Artetxe, Ruder, and Yogatama}]{artetxe-etal-2020-cross}
Mikel Artetxe, Sebastian Ruder, and Dani Yogatama. 2020.
\newblock \href {https://doi.org/10.18653/v1/2020.acl-main.421} {On the cross-lingual transferability of monolingual representations}.
\newblock In \emph{Proceedings of the 58th Annual Meeting of the Association for Computational Linguistics}, pages 4623--4637, Online. Association for Computational Linguistics.

\bibitem[{Bethard(2022)}]{bethard2022randomseeds}
Steven Bethard. 2022.
\newblock \href {https://arxiv.org/abs/2210.13393} {{We need to talk about random seeds}}.
\newblock \emph{Preprint}, arXiv:2210.13393.

\bibitem[{Blackwell et~al.(2025)Blackwell, Barry, and Cohn}]{blackwell2025towardsrepro}
Robert~E. Blackwell, Jon Barry, and Anthony~G. Cohn. 2025.
\newblock \href {https://arxiv.org/abs/2410.03492} {{Towards Reproducible LLM Evaluation: Quantifying Uncertainty in LLM Benchmark Scores}}.
\newblock \emph{Preprint}, arXiv:2410.03492.

\bibitem[{Chen et~al.(2025)Chen, Liu, Da, Chen, Papalexakis, and Wei}]{chen2025uq}
Tiejin Chen, Xiaoou Liu, Longchao Da, Jia Chen, Vagelis Papalexakis, and Hua Wei. 2025.
\newblock \href {https://arxiv.org/abs/2502.16820} {{Uncertainty Quantification of Large Language Models through Multi-Dimensional Responses}}.
\newblock \emph{Preprint}, arXiv:2502.16820.

\bibitem[{Chen et~al.(2024)Chen, Hong, and Madireddy}]{chen2024questionrephrasing}
Zizhang Chen, Pengyu Hong, and Sandeep Madireddy. 2024.
\newblock \href {https://arxiv.org/abs/2408.03732} {{Question Rephrasing for Quantifying Uncertainty in Large Language Models: Applications in Molecular Chemistry Tasks}}.
\newblock \emph{Preprint}, arXiv:2408.03732.

\bibitem[{Conneau et~al.(2020)Conneau, Khandelwal, Goyal, Chaudhary, Wenzek, Guzm{\'a}n, Grave, Ott, Zettlemoyer, and Stoyanov}]{conneau-etal-2020-unsupervised}
Alexis Conneau, Kartikay Khandelwal, Naman Goyal, Vishrav Chaudhary, Guillaume Wenzek, Francisco Guzm{\'a}n, Edouard Grave, Myle Ott, Luke Zettlemoyer, and Veselin Stoyanov. 2020.
\newblock \href {https://doi.org/10.18653/v1/2020.acl-main.747} {Unsupervised cross-lingual representation learning at scale}.
\newblock In \emph{Proceedings of the 58th Annual Meeting of the Association for Computational Linguistics}, pages 8440--8451, Online. Association for Computational Linguistics.

\bibitem[{Da et~al.(2025)Da, Liu, Dai, Cheng, Wang, and Wei}]{da2025understanding}
Longchao Da, Xiaoou Liu, Jiaxin Dai, Lu~Cheng, Yaqing Wang, and Hua Wei. 2025.
\newblock \href {https://openreview.net/forum?id=p4wZfBFgyI} {{Understanding the Uncertainty of {LLM} Explanations: A Perspective Based on Reasoning Topology}}.
\newblock In \emph{Second Conference on Language Modeling}.

\bibitem[{Dang et~al.(2024)Dang, Singh, D'souza, Ahmadian, Salamanca, Smith, Peppin, Hong, Govindassamy, Zhao, Kublik, Amer, Aryabumi, Campos, Tan, Kocmi, Strub, Grinsztajn, Flet-Berliac, Locatelli, Lin, Talupuru, Venkitesh, Cairuz, Yang, Chung, Ko, Shi, Shukayev, Bae, Piktus, Castagné, Cruz-Salinas, Kim, Crawhall-Stein, Morisot, Roy, Blunsom, Zhang, Gomez, Frosst, Fadaee, Ermis, Üstün, and Hooker}]{dang2024ayaexpansecombiningresearch}
John Dang, Shivalika Singh, Daniel D'souza, Arash Ahmadian, Alejandro Salamanca, Madeline Smith, Aidan Peppin, Sungjin Hong, Manoj Govindassamy, Terrence Zhao, Sandra Kublik, Meor Amer, Viraat Aryabumi, Jon~Ander Campos, Yi-Chern Tan, Tom Kocmi, Florian Strub, Nathan Grinsztajn, Yannis Flet-Berliac, Acyr Locatelli, Hangyu Lin, Dwarak Talupuru, Bharat Venkitesh, David Cairuz, Bowen Yang, Tim Chung, Wei-Yin Ko, Sylvie~Shang Shi, Amir Shukayev, Sammie Bae, Aleksandra Piktus, Roman Castagné, Felipe Cruz-Salinas, Eddie Kim, Lucas Crawhall-Stein, Adrien Morisot, Sudip Roy, Phil Blunsom, Ivan Zhang, Aidan Gomez, Nick Frosst, Marzieh Fadaee, Beyza Ermis, Ahmet Üstün, and Sara Hooker. 2024.
\newblock \href {https://arxiv.org/abs/2412.04261} {{Aya Expanse: Combining Research Breakthroughs for a New Multilingual Frontier}}.
\newblock \emph{Preprint}, arXiv:2412.04261.

\bibitem[{Dehghani et~al.(2021)Dehghani, Tay, Gritsenko, Zhao, Houlsby, Diaz, Metzler, and Vinyals}]{benchmarklottery2021}
Mostafa Dehghani, Yi~Tay, Alexey~A. Gritsenko, Zhe Zhao, Neil Houlsby, Fernando Diaz, Donald Metzler, and Oriol Vinyals. 2021.
\newblock \href {https://arxiv.org/abs/2107.07002} {{The Benchmark Lottery}}.
\newblock \emph{Preprint}, arXiv:2107.07002.

\bibitem[{Dem{\v s}ar(2006)}]{demsar2006statcompmultiple}
Janez Dem{\v s}ar. 2006.
\newblock \href {http://jmlr.org/papers/v7/demsar06a.html} {{Statistical Comparisons of Classifiers over Multiple Data Sets}}.
\newblock \emph{Journal of Machine Learning Research}, 7(1):1--30.

\bibitem[{Devlin et~al.(2019)Devlin, Chang, Lee, and Toutanova}]{devlin-etal-2019-bert}
Jacob Devlin, Ming-Wei Chang, Kenton Lee, and Kristina Toutanova. 2019.
\newblock \href {https://doi.org/10.18653/v1/N19-1423} {{BERT}: Pre-training of deep bidirectional transformers for language understanding}.
\newblock In \emph{Proceedings of the 2019 Conference of the North {A}merican Chapter of the Association for Computational Linguistics: Human Language Technologies, Volume 1 (Long and Short Papers)}, pages 4171--4186, Minneapolis, Minnesota. Association for Computational Linguistics.

\bibitem[{Dietterich(1998)}]{dietterich1998approxtests}
Thomas~G. Dietterich. 1998.
\newblock \href {https://doi.org/10.1162/089976698300017197} {{Approximate Statistical Tests for Comparing Supervised Classification Learning Algorithms}}.
\newblock \emph{Neural Computation}, 10:1895--1923.

\bibitem[{Dror et~al.(2018)Dror, Baumer, Shlomov, and Reichart}]{dror-etal-2018-hitchhikers}
Rotem Dror, Gili Baumer, Segev Shlomov, and Roi Reichart. 2018.
\newblock \href {https://doi.org/10.18653/v1/P18-1128} {The hitchhiker{'}s guide to testing statistical significance in natural language processing}.
\newblock In \emph{Proceedings of the 56th Annual Meeting of the Association for Computational Linguistics (Volume 1: Long Papers)}, pages 1383--1392, Melbourne, Australia. Association for Computational Linguistics.

\bibitem[{Efron(1979)}]{efron1979bootstrap}
B~Efron. 1979.
\newblock {Bootstrap Methods: Another Look at the Jackknife}.
\newblock \emph{The Annals of Statistics}, 7(1):1--26.

\bibitem[{Efron(1981)}]{efron1981nonparametric}
Bradley Efron. 1981.
\newblock {Nonparametric estimates of standard error: The jackknife, the bootstrap and other methods}.
\newblock \emph{Biometrika}, 68(3):589--599.

\bibitem[{Gao et~al.(2024)Gao, Tow, Abbasi, Biderman, Black, DiPofi, Foster, Golding, Hsu, Le~Noac'h, Li, McDonell, Muennighoff, Ociepa, Phang, Reynolds, Schoelkopf, Skowron, Sutawika, Tang, Thite, Wang, Wang, and Zou}]{eval-harness}
Leo Gao, Jonathan Tow, Baber Abbasi, Stella Biderman, Sid Black, Anthony DiPofi, Charles Foster, Laurence Golding, Jeffrey Hsu, Alain Le~Noac'h, Haonan Li, Kyle McDonell, Niklas Muennighoff, Chris Ociepa, Jason Phang, Laria Reynolds, Hailey Schoelkopf, Aviya Skowron, Lintang Sutawika, Eric Tang, Anish Thite, Ben Wang, Kevin Wang, and Andy Zou. 2024.
\newblock \href {https://doi.org/10.5281/zenodo.12608602} {{A framework for few-shot language model evaluation}}.

\bibitem[{Gelman(2005)}]{gelman2005anova}
Andrew Gelman. 2005.
\newblock \href {https://doi.org/10.1214/009053604000001048} {{Analysis of variance: Why it is more important than ever}}.
\newblock \emph{The Annals of Statistics}, 33(1):1--31.

\bibitem[{Gelman(2018)}]{gelman2018failure}
Andrew Gelman. 2018.
\newblock {The failure of null hypothesis significance testing when studying incremental changes, and what to do about it}.
\newblock \emph{{Personality and Social Psychology Bulletin}}, 44(1):16--23.

\bibitem[{Gorman and Bedrick(2019)}]{gorman-bedrick-2019-need}
Kyle Gorman and Steven Bedrick. 2019.
\newblock \href {https://doi.org/10.18653/v1/P19-1267} {We need to talk about standard splits}.
\newblock In \emph{Proceedings of the 57th Annual Meeting of the Association for Computational Linguistics}, pages 2786--2791, Florence, Italy. Association for Computational Linguistics.

\bibitem[{Higgins et~al.(2008)Higgins, Thompson, and Spiegelhalter}]{rema2008}
Julian P.~T. Higgins, Simon~G. Thompson, and David~J. Spiegelhalter. 2008.
\newblock \href {https://doi.org/10.1111/j.1467-985X.2008.00552.x} {{A Re-Evaluation of Random-Effects Meta-Analysis}}.
\newblock \emph{Journal of the Royal Statistical Society Series A: Statistics in Society}, 172(1):137--159.

\bibitem[{Imani et~al.(2023)Imani, Lin, Kargaran, Severini, Jalili~Sabet, Kassner, Ma, Schmid, Martins, Yvon, and Sch{\"u}tze}]{imanigooghari-etal-2023-glot500}
Ayyoob Imani, Peiqin Lin, Amir~Hossein Kargaran, Silvia Severini, Masoud Jalili~Sabet, Nora Kassner, Chunlan Ma, Helmut Schmid, Andr{\'e} Martins, Fran{\c{c}}ois Yvon, and Hinrich Sch{\"u}tze. 2023.
\newblock \href {https://doi.org/10.18653/v1/2023.acl-long.61} {Glot500: Scaling multilingual corpora and language models to 500 languages}.
\newblock In \emph{Proceedings of the 61st Annual Meeting of the Association for Computational Linguistics (Volume 1: Long Papers)}, pages 1082--1117, Toronto, Canada. Association for Computational Linguistics.

\bibitem[{Kocmi et~al.(2021)Kocmi, Federmann, Grundkiewicz, Junczys-Dowmunt, Matsushita, and Menezes}]{kocmi-etal-2021-ship}
Tom Kocmi, Christian Federmann, Roman Grundkiewicz, Marcin Junczys-Dowmunt, Hitokazu Matsushita, and Arul Menezes. 2021.
\newblock \href {https://aclanthology.org/2021.wmt-1.57/} {To ship or not to ship: An extensive evaluation of automatic metrics for machine translation}.
\newblock In \emph{Proceedings of the Sixth Conference on Machine Translation}, pages 478--494, Online. Association for Computational Linguistics.

\bibitem[{Kodner et~al.(2023)Kodner, Payne, Khalifa, and Liu}]{kodner-etal-2023-morphological}
Jordan Kodner, Sarah Payne, Salam Khalifa, and Zoey Liu. 2023.
\newblock \href {https://doi.org/10.18653/v1/2023.acl-long.335} {Morphological inflection: A reality check}.
\newblock In \emph{Proceedings of the 61st Annual Meeting of the Association for Computational Linguistics (Volume 1: Long Papers)}, pages 6082--6101, Toronto, Canada. Association for Computational Linguistics.

\bibitem[{Koehn(2004)}]{koehn-2004-statistical}
Philipp Koehn. 2004.
\newblock \href {https://aclanthology.org/W04-3250/} {Statistical significance tests for machine translation evaluation}.
\newblock In \emph{Proceedings of the 2004 Conference on Empirical Methods in Natural Language Processing}, pages 388--395, Barcelona, Spain. Association for Computational Linguistics.

\bibitem[{Kwon et~al.(2023)Kwon, Li, Zhuang, Sheng, Zheng, Yu, Gonzalez, Zhang, and Stoica}]{kwon2023efficient}
Woosuk Kwon, Zhuohan Li, Siyuan Zhuang, Ying Sheng, Lianmin Zheng, Cody~Hao Yu, Joseph~E. Gonzalez, Hao Zhang, and Ion Stoica. 2023.
\newblock {Efficient Memory Management for Large Language Model Serving with PagedAttention}.
\newblock In \emph{Proceedings of the ACM SIGOPS 29th Symposium on Operating Systems Principles}.

\bibitem[{Liu and Dorr(2024)}]{liu-dorr-2024-effect}
Zoey Liu and Bonnie Dorr. 2024.
\newblock \href {https://doi.org/10.18653/v1/2024.naacl-long.157} {The effect of data partitioning strategy on model generalizability: A case study of morphological segmentation}.
\newblock In \emph{Proceedings of the 2024 Conference of the North American Chapter of the Association for Computational Linguistics: Human Language Technologies (Volume 1: Long Papers)}, pages 2851--2864, Mexico City, Mexico. Association for Computational Linguistics.

\bibitem[{Lo et~al.(2023)Lo, Knowles, and Goutte}]{lo-etal-2023-beyond}
Chi-kiu Lo, Rebecca Knowles, and Cyril Goutte. 2023.
\newblock \href {https://aclanthology.org/2023.mtsummit-research.16/} {Beyond correlation: Making sense of the score differences of new {MT} evaluation metrics}.
\newblock In \emph{Proceedings of Machine Translation Summit XIX, Vol. 1: Research Track}, pages 186--199, Macau SAR, China. Asia-Pacific Association for Machine Translation.

\bibitem[{Longjohn et~al.(2025)Longjohn, Gopalan, and Casleton}]{longjohngopalan2025statisticaluq}
Rachel Longjohn, Giri Gopalan, and Emily Casleton. 2025.
\newblock \href {https://arxiv.org/abs/2501.04234} {{Statistical Uncertainty Quantification for Aggregate Performance Metrics in Machine Learning Benchmarks}}.
\newblock \emph{Preprint}, arXiv:2501.04234.

\bibitem[{Nikitin et~al.(2024)Nikitin, Kossen, Gal, and Marttinen}]{nikitin2024kernellanguageentropy}
Alexander~V Nikitin, Jannik Kossen, Yarin Gal, and Pekka Marttinen. 2024.
\newblock \href {https://openreview.net/forum?id=j2wCrWmgMX} {{Kernel Language Entropy: Fine-grained Uncertainty Quantification for {LLM}s from Semantic Similarities}}.
\newblock In \emph{The Thirty-eighth Annual Conference on Neural Information Processing Systems}.

\bibitem[{{NLLB Team} et~al.(2024){NLLB Team}, Costa-juss{\`a}, Cross, {\c{C}}elebi, Elbayad, Heafield, Heffernan, Kalbassi, Lam, Licht, Maillard, Sun, Wang, Wenzek, Youngblood, Akula, Barrault, Gonzalez, Hansanti, Hoffman, Jarrett, Sadagopan, Rowe, Spruit, Tran, Andrews, Ayan, Bhosale, Edunov, Fan, Gao, Goswami, Guzm{\'a}n, Koehn, Mourachko, Ropers, Saleem, Schwenk, and Wang}]{nllb-24}
{NLLB Team}, Marta~R. Costa-juss{\`a}, James Cross, Onur {\c{C}}elebi, Maha Elbayad, Kenneth Heafield, Kevin Heffernan, Elahe Kalbassi, Janice Lam, Daniel Licht, Jean Maillard, Anna Sun, Skyler Wang, Guillaume Wenzek, Al~Youngblood, Bapi Akula, Loic Barrault, Gabriel~Mejia Gonzalez, Prangthip Hansanti, John Hoffman, Semarley Jarrett, Kaushik~Ram Sadagopan, Dirk Rowe, Shannon Spruit, Chau Tran, Pierre Andrews, Necip~Fazil Ayan, Shruti Bhosale, Sergey Edunov, Angela Fan, Cynthia Gao, Vedanuj Goswami, Francisco Guzm{\'a}n, Philipp Koehn, Alexandre Mourachko, Christophe Ropers, Safiyyah Saleem, Holger Schwenk, and Jeff Wang. 2024.
\newblock \href {https://doi.org/10.1038/s41586-024-07335-x} {{Scaling neural machine translation to 200 languages}}.
\newblock \emph{Nature}, 630(8018):841--846.

\bibitem[{Noreen(1989)}]{noreen1989computer}
Eric~W Noreen. 1989.
\newblock \emph{{Computer-intensive methods for testing hypotheses}}.
\newblock John Wiley \& Sons.

\bibitem[{Palen-Michel et~al.(2025)Palen-Michel, Pickering, Kruse, S{\"a}lev{\"a}, and Lignos}]{palen-michel-etal-2025-openner}
Chester Palen-Michel, Maxwell Pickering, Maya Kruse, Jonne S{\"a}lev{\"a}, and Constantine Lignos. 2025.
\newblock \href {https://doi.org/10.18653/v1/2025.emnlp-main.1708} {{O}pen{NER} 1.0: Standardized open-access named entity recognition datasets in 50+ languages}.
\newblock In \emph{Proceedings of the 2025 Conference on Empirical Methods in Natural Language Processing}, pages 33637--33662, Suzhou, China. Association for Computational Linguistics.

\bibitem[{Pan et~al.(2017)Pan, Zhang, May, Nothman, Knight, and Ji}]{pan-etal-2017-cross}
Xiaoman Pan, Boliang Zhang, Jonathan May, Joel Nothman, Kevin Knight, and Heng Ji. 2017.
\newblock \href {https://doi.org/10.18653/v1/P17-1178} {Cross-lingual name tagging and linking for 282 languages}.
\newblock In \emph{Proceedings of the 55th Annual Meeting of the Association for Computational Linguistics (Volume 1: Long Papers)}, pages 1946--1958, Vancouver, Canada. Association for Computational Linguistics.

\bibitem[{Papineni et~al.(2002)Papineni, Roukos, Ward, and Zhu}]{papineni-etal-2002-bleu}
Kishore Papineni, Salim Roukos, Todd Ward, and Wei-Jing Zhu. 2002.
\newblock \href {https://doi.org/10.3115/1073083.1073135} {{B}leu: a method for automatic evaluation of machine translation}.
\newblock In \emph{Proceedings of the 40th Annual Meeting of the Association for Computational Linguistics}, pages 311--318, Philadelphia, Pennsylvania, USA. Association for Computational Linguistics.

\bibitem[{Popovi{\'c}(2015)}]{popovic-2015-chrf}
Maja Popovi{\'c}. 2015.
\newblock \href {https://doi.org/10.18653/v1/W15-3049} {chr{F}: character n-gram {F}-score for automatic {MT} evaluation}.
\newblock In \emph{Proceedings of the Tenth Workshop on Statistical Machine Translation}, pages 392--395, Lisbon, Portugal. Association for Computational Linguistics.

\bibitem[{Post(2018)}]{post-2018-call}
Matt Post. 2018.
\newblock \href {https://doi.org/10.18653/v1/W18-6319} {A call for clarity in reporting {BLEU} scores}.
\newblock In \emph{Proceedings of the Third Conference on Machine Translation: Research Papers}, pages 186--191, Brussels, Belgium. Association for Computational Linguistics.

\bibitem[{Rajpurkar et~al.(2016)Rajpurkar, Zhang, Lopyrev, and Liang}]{rajpurkar-etal-2016-squad}
Pranav Rajpurkar, Jian Zhang, Konstantin Lopyrev, and Percy Liang. 2016.
\newblock \href {https://doi.org/10.18653/v1/D16-1264} {{SQ}u{AD}: 100,000+ questions for machine comprehension of text}.
\newblock In \emph{Proceedings of the 2016 Conference on Empirical Methods in Natural Language Processing}, pages 2383--2392, Austin, Texas. Association for Computational Linguistics.

\bibitem[{Rei et~al.(2020)Rei, Stewart, Farinha, and Lavie}]{rei-etal-2020-comet}
Ricardo Rei, Craig Stewart, Ana~C Farinha, and Alon Lavie. 2020.
\newblock \href {https://doi.org/10.18653/v1/2020.emnlp-main.213} {{COMET}: A neural framework for {MT} evaluation}.
\newblock In \emph{Proceedings of the 2020 Conference on Empirical Methods in Natural Language Processing (EMNLP)}, pages 2685--2702, Online. Association for Computational Linguistics.

\bibitem[{Reimers and Gurevych(2017)}]{reimers-gurevych-2017-reporting}
Nils Reimers and Iryna Gurevych. 2017.
\newblock \href {https://doi.org/10.18653/v1/D17-1035} {Reporting score distributions makes a difference: Performance study of {LSTM}-networks for sequence tagging}.
\newblock In \emph{Proceedings of the 2017 Conference on Empirical Methods in Natural Language Processing}, pages 338--348, Copenhagen, Denmark. Association for Computational Linguistics.

\bibitem[{Rubin(1981)}]{rubin1981bayesboot}
Donald~B. Rubin. 1981.
\newblock \href {https://doi.org/10.1214/aos/1176345338} {{The Bayesian Bootstrap}}.
\newblock \emph{The Annals of Statistics}, 9(1):130--134.

\bibitem[{S{\o}gaard et~al.(2021)S{\o}gaard, Ebert, Bastings, and Filippova}]{sogaard-etal-2021-need}
Anders S{\o}gaard, Sebastian Ebert, Jasmijn Bastings, and Katja Filippova. 2021.
\newblock \href {https://doi.org/10.18653/v1/2021.eacl-main.156} {We need to talk about random splits}.
\newblock In \emph{Proceedings of the 16th Conference of the European Chapter of the Association for Computational Linguistics: Main Volume}, pages 1823--1832, Online. Association for Computational Linguistics.

\bibitem[{Tatiana and Valentin(2021)}]{tatiana2021not}
Shavrina Tatiana and Malykh Valentin. 2021.
\newblock \href {https://arxiv.org/abs/2112.01342} {{How not to Lie with a Benchmark: Rearranging NLP Leaderboards}}.
\newblock \emph{Preprint}, arXiv:2112.01342.

\bibitem[{Ulmer et~al.(2022)Ulmer, Bassignana, M{\"u}ller-Eberstein, Varab, Zhang, van~der Goot, Hardmeier, and Plank}]{ulmer-etal-2022-experimental}
Dennis Ulmer, Elisa Bassignana, Max M{\"u}ller-Eberstein, Daniel Varab, Mike Zhang, Rob van~der Goot, Christian Hardmeier, and Barbara Plank. 2022.
\newblock \href {https://doi.org/10.18653/v1/2022.findings-emnlp.196} {Experimental standards for deep learning in natural language processing research}.
\newblock In \emph{Findings of the Association for Computational Linguistics: EMNLP 2022}, pages 2673--2692, Abu Dhabi, United Arab Emirates. Association for Computational Linguistics.

\bibitem[{Vasishth and Gelman(2021)}]{vasishth2021embrace}
Shravan Vasishth and Andrew Gelman. 2021.
\newblock \href {https://doi.org/10.1515/ling-2019-0051} {{How to embrace variation and accept uncertainty in linguistic and psycholinguistic data analysis}}.
\newblock \emph{Linguistics}, 59:1311--1342.

\bibitem[{Wagner et~al.(2024)Wagner, Desmond, Nair, Ashktorab, Daly, Pan, Cooper, Johnson, and Geyer}]{wagner2024blackbox}
Nico Wagner, Michael Desmond, Rahul Nair, Zahra Ashktorab, Elizabeth~M. Daly, Qian Pan, Martín~Santillán Cooper, James~M. Johnson, and Werner Geyer. 2024.
\newblock \href {https://arxiv.org/abs/2410.11594} {{Black-box Uncertainty Quantification Method for LLM-as-a-Judge}}.
\newblock \emph{Preprint}, arXiv:2410.11594.

\bibitem[{Xiang et~al.(2022)Xiang, Li, Liu, Liu, Huang, Lian, and Shi}]{xiang-etal-2022-investigating}
Jiannan Xiang, Huayang Li, Yahui Liu, Lemao Liu, Guoping Huang, Defu Lian, and Shuming Shi. 2022.
\newblock \href {https://doi.org/10.18653/v1/2022.findings-acl.14} {Investigating data variance in evaluations of automatic machine translation metrics}.
\newblock In \emph{Findings of the Association for Computational Linguistics: ACL 2022}, pages 150--157, Dublin, Ireland. Association for Computational Linguistics.

\bibitem[{Yang et~al.(2025)Yang, Yoo, and Lee}]{yang-etal-2025-maqa}
Yongjin Yang, Haneul Yoo, and Hwaran Lee. 2025.
\newblock \href {https://doi.org/10.18653/v1/2025.findings-naacl.325} {{MAQA}: Evaluating uncertainty quantification in {LLM}s regarding data uncertainty}.
\newblock In \emph{Findings of the Association for Computational Linguistics: NAACL 2025}, pages 5846--5863, Albuquerque, New Mexico. Association for Computational Linguistics.

\bibitem[{Ye et~al.(2024)Ye, Yang, Pang, Wang, Wong, Yilmaz, Shi, and Tu}]{ye2024benchmarking}
Fanghua Ye, Mingming Yang, Jianhui Pang, Longyue Wang, Derek~F. Wong, Emine Yilmaz, Shuming Shi, and Zhaopeng Tu. 2024.
\newblock \href {https://doi.org/10.52202/079017-0491} {{Benchmarking LLMs via Uncertainty Quantification}}.
\newblock In \emph{Advances in Neural Information Processing Systems}, volume~37, pages 15356--15385. Curran Associates, Inc.

\end{thebibliography}

\appendix

\section{Experimental settings}

\subsection{Question answering}

We evaluate four LLMs on all XQuAD subsets: \texttt{aya-expanse-8B} \citep{dang2024ayaexpansecombiningresearch}, \texttt{TowerInstruct-Mistral-7B-v0.2} \citep{alves2024tower}, Google's \texttt{gemma2-9b} and finally \texttt{Clarus-7B-v0.3}. All models are taken from the Open LLM Leaderboard\footnote{\url{https://huggingface.co/spaces/open-llm-leaderboard/open_llm_leaderboard}} from HuggingFace.
Details of the experimental settings are in the Appendix.
All experiments are run using the \texttt{lm-evaluation-harness} library \citep{eval-harness} on A40 40GB GPUs using vLLM as an inference back-end \citep{kwon2023efficient}.
We evaluate using token-level F1 score, using the standard \texttt{lm-evaluation-harness} implementation. We do not consider exact match as it is less robust to random deviations.

\subsection{Machine translation}

For each language we create two language pairs: XX$\rightarrow$EN and EN$\rightarrow$XX.
All of our experiments use use the \texttt{devtest} split of FLORES-200 which contains approximately 1,000 sentences per language. 
All experiments are run using the \texttt{lm-evaluation-harness} library \citep{eval-harness} on A40 40GB GPUs.
We evaluate using BLEU \citep{papineni-etal-2002-bleu},  ChrF++ \citep{popovic-2015-chrf} and COMET \citep{rei-etal-2020-comet}.

\section{Datasets used}

\subsection{XQuAD}
\label{xquad-appendix}

XQuAD \citep{artetxe-etal-2020-cross} is a multilingual extension of the SQuAD v1.1 question answering benchmark \citep{rajpurkar-etal-2016-squad}.
XQuAD includes translations 1,190 questions, originally in English, into Arabic, German, Greek, Spanish, Hindi, Romanian, Russian, Thai, Turkish, Vietnamese, and Mandarin Chinese.

\subsection{OpenNER 1.0}

Our analysis is based on the OpenNER 1.0 multilingual NER benchmark \citep{palen-michel-etal-2025-openner} which consists of 61 unique datasets covering a total of 51 languages. 
The authors experiment with three pretrained language models (PLM): mBERT \citep{devlin-etal-2019-bert}, XLM-R \citep{conneau-etal-2020-unsupervised}, and Glot500-m \citep{imanigooghari-etal-2023-glot500}.
Each PLM is finetuned using two experimental conditions per language.
In the ``individual'' condition, each PLM is finetuned separately on the train split of each of the 61 datasets.
In the ``concatenated'' setting, the train splits of all 61 datasets are concatenated\footnote{Due to computational considerations, before concatenation, the highest-resource datasets are downsampled.} together before finetuning each of the three PLMs on it.
The authors run each experiment 10 times using different seeds (values 42--51).

We use model outputs from the original authors with their permission.
In addition to the original results, we construct $B=100$ bootstrap replicated data sets and reuse the same predictions from the original test set. 
This gives us a total of $R = SB = 1000$ replicated scores for each of the 61 datasets.

\section{Additional Tables and Figures}
Additional tables follow.

\subsection{Variance components}

\begin{table*}[p]
    \centering
    \small
    \begin{tabular}{lrrr}
\toprule
& \multicolumn{2}{c}{Aya Expanse 8B} & TowerInstruct 7B \\
\cmidrule(lr){2-3} \cmidrule(lr){4-4}
Task & TowerInstruct 7B & Clarus 7B & Clarus 7B \\
\midrule
\multicolumn{4}{l}{BLEU} \\
\midrule
de-en & \textbf{0.40 $ \pm $ 1.34} & 12.31 $ \pm $ 1.14 & 11.91 $ \pm $ 1.46 \\
en-de & \textbf{-1.13 $ \pm $ 1.13} & 12.48 $ \pm $ 0.88 & 13.61 $ \pm $ 1.03 \\
en-es & \textbf{-0.35 $ \pm $ 0.79} & 8.64 $ \pm $ 0.76 & 8.99 $ \pm $ 0.90 \\
en-fr & \textbf{1.01 $ \pm $ 1.18} & 15.32 $ \pm $ 1.32 & 14.30 $ \pm $ 1.37 \\
en-it & \textbf{1.17 $ \pm $ 0.74} & 11.96 $ \pm $ 0.81 & 10.79 $ \pm $ 0.80 \\
en-ko & 1.88 $ \pm $ 0.49 & 5.15 $ \pm $ 0.45 & 3.27 $ \pm $ 0.35 \\
en-nl & \textbf{0.40 $ \pm $ 0.90} & 10.05 $ \pm $ 0.72 & 9.65 $ \pm $ 0.80 \\
en-pt & -7.16 $ \pm $ 1.03 & 12.85 $ \pm $ 1.32 & 20.01 $ \pm $ 1.35 \\
en-ru & \textbf{-0.25 $ \pm $ 0.94} & 12.31 $ \pm $ 0.76 & 12.56 $ \pm $ 0.82 \\
es-en & \textbf{-1.29 $ \pm $ 1.06} & 9.41 $ \pm $ 0.93 & 10.70 $ \pm $ 0.93 \\
fr-en & \textbf{-1.54 $ \pm $ 1.33} & 14.05 $ \pm $ 1.13 & 15.59 $ \pm $ 1.37 \\
it-en & 2.47 $ \pm $ 1.10 & 11.16 $ \pm $ 0.98 & 8.69 $ \pm $ 1.21 \\
ko-en & -4.15 $ \pm $ 0.96 & 6.51 $ \pm $ 0.81 & 10.66 $ \pm $ 0.95 \\
nl-en & \textbf{-1.78 $ \pm $ 1.01} & 10.14 $ \pm $ 0.90 & 11.92 $ \pm $ 1.09 \\
pt-en & -5.67 $ \pm $ 1.07 & 18.38 $ \pm $ 1.26 & 24.05 $ \pm $ 1.24 \\
ru-en & -2.86 $ \pm $ 0.95 & 10.01 $ \pm $ 1.06 & 12.86 $ \pm $ 1.04 \\
\midrule
\multicolumn{4}{l}{ChrF++} \\
\midrule
de-en & -2.08 $ \pm $ 0.76 & 6.53 $ \pm $ 0.94 & 8.62 $ \pm $ 0.94 \\
en-de & -2.03 $ \pm $ 0.72 & 12.35 $ \pm $ 0.82 & 14.37 $ \pm $ 0.80 \\
en-es & \textbf{-0.97 $ \pm $ 0.61} & 6.48 $ \pm $ 0.69 & 7.45 $ \pm $ 0.69 \\
en-fr & \textbf{0.04 $ \pm $ 0.82} & 10.40 $ \pm $ 0.91 & 10.36 $ \pm $ 0.91 \\
en-it & \textbf{0.20 $ \pm $ 0.55} & 11.09 $ \pm $ 0.67 & 10.89 $ \pm $ 0.61 \\
en-ko & \textbf{-0.20 $ \pm $ 0.62} & 14.27 $ \pm $ 0.57 & 14.47 $ \pm $ 0.53 \\
en-nl & \textbf{-0.44 $ \pm $ 0.67} & 11.50 $ \pm $ 0.74 & 11.94 $ \pm $ 0.74 \\
en-pt & -6.49 $ \pm $ 0.71 & 7.78 $ \pm $ 0.86 & 14.28 $ \pm $ 0.80 \\
en-ru & \textbf{-1.36 $ \pm $ 0.78} & 15.33 $ \pm $ 0.80 & 16.69 $ \pm $ 0.78 \\
es-en & -3.99 $ \pm $ 0.79 & 5.45 $ \pm $ 0.95 & 9.43 $ \pm $ 0.79 \\
fr-en & -3.74 $ \pm $ 0.88 & 7.09 $ \pm $ 0.92 & 10.83 $ \pm $ 0.93 \\
it-en & \textbf{-0.79 $ \pm $ 0.76} & 7.03 $ \pm $ 0.94 & 7.82 $ \pm $ 0.95 \\
ko-en & -4.61 $ \pm $ 0.63 & 5.52 $ \pm $ 0.68 & 10.13 $ \pm $ 0.70 \\
nl-en & -3.74 $ \pm $ 0.74 & 5.37 $ \pm $ 1.03 & 9.11 $ \pm $ 1.03 \\
pt-en & -6.04 $ \pm $ 0.73 & 9.46 $ \pm $ 1.15 & 15.50 $ \pm $ 1.07 \\
ru-en & -4.11 $ \pm $ 0.69 & 5.46 $ \pm $ 1.05 & 9.58 $ \pm $ 0.99 \\
\midrule
\multicolumn{4}{l}{COMET} \\
\midrule
de-en & \textbf{0.92 $ \pm $ 0.50} & 6.60 $ \pm $ 0.79 & 5.68 $ \pm $ 0.86 \\
en-de & 1.19 $ \pm $ 0.57 & 16.89 $ \pm $ 0.93 & 15.70 $ \pm $ 0.89 \\
en-es & 1.19 $ \pm $ 0.59 & 10.46 $ \pm $ 0.92 & 9.27 $ \pm $ 0.89 \\
en-fr & \textbf{0.35 $ \pm $ 0.48} & 10.86 $ \pm $ 0.71 & 10.50 $ \pm $ 0.73 \\
en-it & \textbf{0.83 $ \pm $ 0.45} & 14.74 $ \pm $ 0.78 & 13.90 $ \pm $ 0.77 \\
en-ko & 5.16 $ \pm $ 0.73 & 24.62 $ \pm $ 0.94 & 19.45 $ \pm $ 0.89 \\
en-nl & 1.26 $ \pm $ 0.53 & 19.53 $ \pm $ 0.78 & 18.27 $ \pm $ 0.83 \\
en-pt & -1.85 $ \pm $ 0.47 & 10.32 $ \pm $ 0.86 & 12.17 $ \pm $ 0.85 \\
en-ru & \textbf{-0.53 $ \pm $ 0.54} & 17.45 $ \pm $ 0.87 & 17.98 $ \pm $ 0.85 \\
es-en & \textbf{0.68 $ \pm $ 0.62} & 7.64 $ \pm $ 0.81 & 6.97 $ \pm $ 0.84 \\
fr-en & \textbf{0.31 $ \pm $ 0.55} & 7.79 $ \pm $ 0.83 & 7.48 $ \pm $ 0.84 \\
it-en & 2.57 $ \pm $ 0.72 & 7.76 $ \pm $ 0.71 & 5.19 $ \pm $ 0.89 \\
ko-en & \textbf{-0.78 $ \pm $ 0.41} & 6.51 $ \pm $ 0.59 & 7.29 $ \pm $ 0.61 \\
nl-en & \textbf{-0.68 $ \pm $ 0.57} & 7.58 $ \pm $ 0.89 & 8.26 $ \pm $ 0.95 \\
pt-en & -1.81 $ \pm $ 0.40 & 9.40 $ \pm $ 1.07 & 11.21 $ \pm $ 1.04 \\
ru-en & \textbf{-0.57 $ \pm $ 0.40} & 6.33 $ \pm $ 0.73 & 6.90 $ \pm $ 0.70 \\
\bottomrule
\end{tabular}

    \caption{Pairwise differences between models on NMT experiments, with uncertainty quantified using bootstrap resampling and multiple seeds. The error bar corresponds $\pm$ one standard error. Nonsignificant differences are indicated in boldface.} %The highest score is bolded when the difference is more than 2 standard errors from zero.}
    \label{tab:flores_results_uncert_seedboot}
\end{table*}

\paragraph{Summary variance component tables} 

Table~\ref{tab:flores_varcomp_table} shows a detailed summary of variance components for NMT experimets.

\begin{table*}[p]
    \centering
    \small
    \begin{tabular}{lrrrr}
\toprule
\multicolumn{5}{c}{BLEU} \\
\midrule
& Mean & SD & Min & Max \\
\midrule
\textit{Between-language total ($\nu$)} \\
\midrule
Aya Expanse 8B & 7.11 & - & - & - \\
TowerInstruct 7B & 8.49 & - & - & - \\
Clarus 7B & 4.50 & - & - & - \\
\midrule
\textit{Within-language total ($\eta_l$)} \\
\midrule
Aya Expanse 8B & 0.64 & 0.10 & 0.40 & 0.80 \\
TowerInstruct 7B & 0.77 & 0.22 & 0.28 & 1.15 \\
Clarus 7B & 0.70 & 0.23 & 0.22 & 1.12 \\
\midrule
\textit{Seed-to-seed ($\sigma_l$)} \\
\midrule
Aya Expanse 8B & 0.34 & 0.10 & 0.15 & 0.55 \\
TowerInstruct 7B & 0.47 & 0.16 & 0.13 & 0.78 \\
Clarus 7B & 0.48 & 0.20 & 0.14 & 0.92 \\
\midrule
\textit{Boot-to-boot ($\tau_l$)} \\
\midrule
Aya Expanse 8B & 0.53 & 0.08 & 0.34 & 0.67 \\
TowerInstruct 7B & 0.60 & 0.17 & 0.25 & 0.93 \\
Clarus 7B & 0.49 & 0.13 & 0.17 & 0.70 \\
\midrule
\multicolumn{5}{c}{ChrF++} \\
\midrule
\textit{Between-language total ($\nu$)} \\
\midrule
Aya Expanse 8B & 8.05 & - & - & - \\
TowerInstruct 7B & 8.99 & - & - & - \\
Clarus 7B & 9.66 & - & - & - \\
\midrule
\textit{Within-language total ($\eta_l$)} \\
\midrule
Aya Expanse 8B & 0.53 & 0.06 & 0.43 & 0.68 \\
TowerInstruct 7B & 0.48 & 0.08 & 0.33 & 0.64 \\
Clarus 7B & 0.67 & 0.16 & 0.34 & 0.99 \\
\midrule
\textit{Seed-to-seed ($\sigma_l$)} \\
\midrule
Aya Expanse 8B & 0.28 & 0.09 & 0.11 & 0.46 \\
TowerInstruct 7B & 0.25 & 0.08 & 0.08 & 0.42 \\
Clarus 7B & 0.43 & 0.18 & 0.11 & 0.82 \\
\midrule
\textit{Boot-to-boot ($\tau_l$)} \\
\midrule
Aya Expanse 8B & 0.44 & 0.04 & 0.35 & 0.51 \\
TowerInstruct 7B & 0.40 & 0.04 & 0.32 & 0.48 \\
Clarus 7B & 0.50 & 0.06 & 0.32 & 0.58 \\
\midrule
\multicolumn{5}{c}{COMET} \\
\midrule
\textit{Between-language total ($\nu$)} \\
\midrule
Aya Expanse 8B & 1.21 & - & - & - \\
TowerInstruct 7B & 2.44 & - & - & - \\
Clarus 7B & 5.75 & - & - & - \\
\midrule
\textit{Within-language total ($\eta_l$)} \\
\midrule
Aya Expanse 8B & 0.36 & 0.07 & 0.26 & 0.55 \\
TowerInstruct 7B & 0.38 & 0.10 & 0.23 & 0.63 \\
Clarus 7B & 0.74 & 0.12 & 0.53 & 1.01 \\
\midrule
\textit{Seed-to-seed ($\sigma_l$)} \\
\midrule
Aya Expanse 8B & 0.20 & 0.06 & 0.10 & 0.34 \\
TowerInstruct 7B & 0.22 & 0.11 & 0.08 & 0.52 \\
Clarus 7B & 0.50 & 0.15 & 0.27 & 0.87 \\
\midrule
\textit{Boot-to-boot ($\tau_l$)} \\
\midrule
Aya Expanse 8B & 0.30 & 0.06 & 0.22 & 0.46 \\
TowerInstruct 7B & 0.30 & 0.06 & 0.21 & 0.44 \\
Clarus 7B & 0.53 & 0.06 & 0.41 & 0.66 \\
\bottomrule
\end{tabular}
    \caption{Summary of variance components for NMT experiments.}
    \label{tab:flores_varcomp_table}
\end{table*}

\begin{table*}[p!]
    \centering
    \small
    \begin{tabular}{lrrrr}
\toprule
& Mean & SD & Min & Max \\
\midrule
\textit{Between-language total ($\sigma$)} \\
\midrule
Glot500 (concat) & 6.30 & - & - & - \\
Glot500 (ind) & 8.27 & - & - & - \\
mBERT (concat) & 14.38 & - & - & - \\
mBERT (ind) & 9.95 & - & - & - \\
XLM-R (concat) & 6.31 & - & - & - \\
XLM-R (ind) & 7.13 & - & - & - \\
\midrule
\textit{Within-language total ($\eta_l$)} \\
\midrule
Glot500 (concat) & 1.33 & 0.86 & 0.33 & 4.88 \\
Glot500 (ind) & 2.19 & 2.56 & 0.39 & 13.98 \\
mBERT (concat) & 1.70 & 0.99 & 0.24 & 5.42 \\
mBERT (ind) & 1.90 & 2.61 & 0.14 & 20.62 \\
XLM-R (concat) & 1.36 & 0.87 & 0.15 & 5.08 \\
XLM-R (ind) & 1.93 & 3.54 & 0.13 & 28.12 \\
\midrule
\textit{Seed-to-seed ($\sigma_l$)} \\
\midrule
Glot500 (concat) & 0.61 & 0.39 & 0.19 & 2.09 \\
Glot500 (ind) & 1.59 & 2.51 & 0.16 & 13.69 \\
mBERT (concat) & 0.81 & 0.45 & 0.12 & 2.29 \\
mBERT (ind) & 1.15 & 2.55 & 0.06 & 20.35 \\
XLM-R (concat) & 0.64 & 0.42 & 0.08 & 2.46 \\
XLM-R (ind) & 1.24 & 3.53 & 0.04 & 28.12 \\
\midrule
\textit{Boot-to-boot ($\tau_l$)} \\
\midrule
Glot500 (concat) & 1.17 & 0.79 & 0.13 & 4.79 \\
Glot500 (ind) & 1.27 & 0.95 & 0.12 & 5.31 \\
mBERT (concat) & 1.47 & 0.91 & 0.21 & 5.33 \\
mBERT (ind) & 1.36 & 0.86 & 0.13 & 5.20 \\
XLM-R (concat) & 1.18 & 0.79 & 0.13 & 5.00 \\
XLM-R (ind) & 1.23 & 0.84 & 0.12 & 5.22 \\
\bottomrule
\end{tabular}

% \begin{tabular}{lrrrr}
% \toprule
% & Mean & SD & Min & Max \\
% \midrule
% \textit{Between-language total ($\sigma$)} \\
% \midrule
% Glot500 (concat) & 6.23 & - & - & - \\
% Glot500 (ind) & 8.16 & - & - & - \\
% mBERT (concat) & 14.28 & - & - & - \\
% mBERT (ind) & 9.83 & - & - & - \\
% XLM-R (concat) & 6.24 & - & - & - \\
% XLM-R (ind) & 7.04 & - & - & - \\
% \midrule
% \textit{Within-language total ($\eta_l$)} \\
% \midrule
% Glot500 (concat) & 1.76 & 0.93 & 0.51 & 4.24 \\
% Glot500 (ind) & 1.49 & 0.52 & 0.68 & 2.74 \\
% mBERT (concat) & 1.27 & 0.76 & 0.12 & 3.14 \\
% mBERT (ind) & 1.26 & 0.75 & 0.53 & 3.53 \\
% XLM-R (concat) & 1.58 & 1.24 & 0.48 & 5.45 \\
% XLM-R (ind) & 1.47 & 1.11 & 0.37 & 4.80 \\
% \midrule
% \textit{Seed-to-seed ($\sigma_l$)} \\
% \midrule
% Glot500 (concat) & 0.84 & 0.62 & 0.13 & 2.63 \\
% Glot500 (ind) & 0.75 & 0.41 & 0.23 & 1.85 \\
% mBERT (concat) & 0.71 & 0.57 & 0.04 & 2.34 \\
% mBERT (ind) & 0.66 & 0.48 & 0.19 & 2.10 \\
% XLM-R (concat) & 0.64 & 0.56 & 0.15 & 2.40 \\
% XLM-R (ind) & 0.66 & 0.51 & 0.14 & 2.18 \\
% \midrule
% \textit{Boot-to-boot ($\tau_l$)} \\
% \midrule
% Glot500 (concat) & 1.54 & 0.81 & 0.47 & 3.72 \\
% Glot500 (ind) & 1.28 & 0.40 & 0.59 & 2.18 \\
% mBERT (concat) & 1.02 & 0.67 & 0.12 & 2.79 \\
% mBERT (ind) & 1.05 & 0.60 & 0.46 & 2.87 \\
% XLM-R (concat) & 1.13 & 1.09 & 0.40 & 4.77 \\
% XLM-R (ind) & 1.30 & 0.74 & 0.33 & 3.28 \\
% \bottomrule
% \end{tabular}
    \caption{Summary of variance components for NER experiments.}
    \label{tab:ner_varcomp_table_summary}
\end{table*}

\paragraph{Detailed variance component tables} 

Table~\ref{tab:varcomp_table_xquad_full} shows detailed estimates for variance components for QA.
Table~\ref{tab:varcomp_table_flores_full} shows detailed estimates for variance components for MT.
Table~\ref{tab:varcomp_table} shows detailed estimates for variance components for NER.

\begin{table*}[p]
    \centering
    \small
    \begin{tabular}{lrrrrrrrrrrrr}
\toprule
 & \multicolumn{3}{c}{Clarus 7B} & \multicolumn{3}{c}{TowerInstruct 7B} & \multicolumn{3}{c}{Aya Expanse 8B} & \multicolumn{3}{c}{Gemma 2 9B}  \\
\cmidrule(lr){2-4} \cmidrule(lr){5-7} \cmidrule(lr){8-10} \cmidrule(lr){11-13}
 Language & $\sigma$ & $\tau$ & $\eta$ & $\sigma$ & $\tau$ & $\eta$ & $\sigma$ & $\tau$ & $\eta$ & $\sigma$ & $\tau$ & $\eta$  \\
\midrule
Arabic & 0.20 & 0.52 & 0.56 & 0.46 & 0.43 & 0.63 & 0.25 & 1.15 & 1.18 & 0.48 & 0.80 & 0.93 \\
Chinese & 0.36 & 0.55 & 0.66 & 0.52 & 0.63 & 0.82 & 0.65 & 1.28 & 1.43 & 0.51 & 0.98 & 1.10 \\
English & 0.93 & 1.27 & 1.58 & 0.60 & 1.05 & 1.21 & 0.96 & 1.41 & 1.71 & 0.95 & 1.18 & 1.51 \\
German & 0.57 & 0.54 & 0.78 & 0.74 & 0.81 & 1.10 & 0.79 & 1.31 & 1.53 & 0.69 & 1.05 & 1.26 \\
Greek & 0.36 & 0.74 & 0.82 & 0.17 & 0.33 & 0.37 & 0.71 & 1.07 & 1.29 & 0.97 & 0.89 & 1.31 \\
Hindi & 0.32 & 0.77 & 0.83 & 0.12 & 0.15 & 0.19 & 0.89 & 1.34 & 1.61 & 0.72 & 0.96 & 1.20 \\
Romanian & 0.36 & 0.80 & 0.88 & 0.51 & 0.76 & 0.91 & 0.34 & 1.33 & 1.37 & 0.85 & 1.01 & 1.32 \\
Russian & 0.88 & 0.79 & 1.18 & 0.31 & 0.95 & 1.00 & 0.53 & 1.21 & 1.32 & 0.80 & 0.84 & 1.16 \\
Spanish & 1.25 & 1.11 & 1.67 & 0.49 & 0.76 & 0.91 & 0.88 & 1.26 & 1.54 & 1.09 & 0.88 & 1.40 \\
Thai & 0.59 & 0.70 & 0.91 & 0.19 & 0.30 & 0.35 & 0.72 & 1.06 & 1.28 & 0.65 & 0.92 & 1.13 \\
Turkish & 0.16 & 0.37 & 0.40 & 0.40 & 0.41 & 0.57 & 1.23 & 1.21 & 1.73 & 0.67 & 1.00 & 1.20 \\
Vietnamese & 0.34 & 0.24 & 0.42 & 0.42 & 0.52 & 0.67 & 1.18 & 1.19 & 1.67 & 0.51 & 0.89 & 1.03 \\
\bottomrule
\end{tabular}

    \caption{Detailed estimates of the standard deviations due to model-related $\sigma$ and bootstrap-related $\tau$ uncertainty in question answering experiments.}
    \label{tab:varcomp_table_xquad_full}
\end{table*}

\begin{table*}[p]
    \centering
    \small
    \begin{tabular}{lrrrrrr}
\toprule
 & \multicolumn{2}{c}{Aya Expanse 8B} & \multicolumn{2}{c}{TowerInstruct 7B} & \multicolumn{2}{c}{Clarus 7B} \\
\cmidrule(lr){2-3} \cmidrule(lr){4-5} \cmidrule(lr){6-7}
 & $\sigma$ & $\tau$ & $\sigma$ & $\tau$ & $\sigma$ & $\tau$ \\
\midrule
\multicolumn{7}{l}{BLEU} \\
\midrule
German - English & 0.39 & 0.57 & 0.67 & 0.93 & 0.58 & 0.70 \\
English - German & 0.45 & 0.54 & 0.59 & 0.65 & 0.24 & 0.48 \\
English - Spanish & 0.17 & 0.41 & 0.48 & 0.44 & 0.44 & 0.44 \\
English - French & 0.45 & 0.66 & 0.62 & 0.64 & 0.82 & 0.67 \\
English - Italian & 0.24 & 0.47 & 0.18 & 0.49 & 0.46 & 0.41 \\
English - Korean & 0.20 & 0.34 & 0.13 & 0.25 & 0.14 & 0.17 \\
English - Dutch & 0.37 & 0.45 & 0.48 & 0.49 & 0.22 & 0.37 \\
English - Portuguese & 0.23 & 0.67 & 0.40 & 0.65 & 0.92 & 0.65 \\
English - Russian & 0.38 & 0.52 & 0.41 & 0.55 & 0.15 & 0.41 \\
Spanish - English & 0.55 & 0.51 & 0.43 & 0.61 & 0.36 & 0.43 \\
French - English & 0.43 & 0.63 & 0.78 & 0.76 & 0.58 & 0.57 \\
Italian - English & 0.26 & 0.53 & 0.65 & 0.66 & 0.59 & 0.51 \\
Korean - English & 0.30 & 0.50 & 0.51 & 0.57 & 0.34 & 0.47 \\
Dutch - English & 0.15 & 0.55 & 0.59 & 0.60 & 0.52 & 0.47 \\
Portuguese - English & 0.46 & 0.61 & 0.32 & 0.68 & 0.79 & 0.59 \\
Russian - English & 0.38 & 0.55 & 0.25 & 0.60 & 0.58 & 0.57 \\
\midrule
\multicolumn{7}{l}{ChrF++} \\
\midrule
German - English & 0.32 & 0.43 & 0.33 & 0.43 & 0.51 & 0.58 \\
English - German & 0.30 & 0.43 & 0.24 & 0.42 & 0.38 & 0.51 \\
English - Spanish & 0.24 & 0.36 & 0.29 & 0.32 & 0.35 & 0.41 \\
English - French & 0.33 & 0.47 & 0.39 & 0.44 & 0.47 & 0.54 \\
English - Italian & 0.25 & 0.35 & 0.08 & 0.32 & 0.31 & 0.41 \\
English - Korean & 0.26 & 0.38 & 0.18 & 0.38 & 0.11 & 0.32 \\
English - Dutch & 0.28 & 0.37 & 0.30 & 0.37 & 0.39 & 0.42 \\
English - Portuguese & 0.21 & 0.50 & 0.21 & 0.41 & 0.39 & 0.53 \\
English - Russian & 0.31 & 0.45 & 0.34 & 0.41 & 0.19 & 0.53 \\
Spanish - English & 0.46 & 0.50 & 0.11 & 0.40 & 0.44 & 0.51 \\
French - English & 0.34 & 0.51 & 0.42 & 0.48 & 0.40 & 0.56 \\
Italian - English & 0.34 & 0.40 & 0.34 & 0.42 & 0.57 & 0.53 \\
Korean - English & 0.14 & 0.41 & 0.20 & 0.41 & 0.16 & 0.50 \\
Dutch - English & 0.11 & 0.51 & 0.29 & 0.44 & 0.72 & 0.52 \\
Portuguese - English & 0.33 & 0.49 & 0.11 & 0.40 & 0.82 & 0.56 \\
Russian - English & 0.29 & 0.46 & 0.17 & 0.41 & 0.70 & 0.54 \\
\midrule
\multicolumn{7}{l}{COMET} \\
\midrule
German - English & 0.14 & 0.22 & 0.32 & 0.28 & 0.60 & 0.44 \\
English - German & 0.31 & 0.33 & 0.13 & 0.32 & 0.55 & 0.61 \\
English - Spanish & 0.34 & 0.28 & 0.26 & 0.29 & 0.55 & 0.57 \\
English - French & 0.14 & 0.29 & 0.22 & 0.29 & 0.27 & 0.57 \\
English - Italian & 0.10 & 0.31 & 0.12 & 0.28 & 0.45 & 0.54 \\
English - Korean & 0.31 & 0.46 & 0.18 & 0.44 & 0.42 & 0.62 \\
English - Dutch & 0.15 & 0.29 & 0.24 & 0.34 & 0.42 & 0.57 \\
English - Portuguese & 0.19 & 0.30 & 0.21 & 0.23 & 0.59 & 0.54 \\
English - Russian & 0.25 & 0.33 & 0.17 & 0.30 & 0.40 & 0.66 \\
Spanish - English & 0.24 & 0.33 & 0.25 & 0.38 & 0.49 & 0.50 \\
French - English & 0.19 & 0.31 & 0.22 & 0.34 & 0.55 & 0.51 \\
Italian - English & 0.23 & 0.25 & 0.52 & 0.36 & 0.39 & 0.49 \\
Korean - English & 0.15 & 0.22 & 0.18 & 0.25 & 0.34 & 0.41 \\
Dutch - English & 0.11 & 0.30 & 0.36 & 0.30 & 0.69 & 0.46 \\
Portuguese - English & 0.20 & 0.26 & 0.08 & 0.21 & 0.87 & 0.51 \\
Russian - English & 0.19 & 0.25 & 0.11 & 0.22 & 0.48 & 0.45 \\
\bottomrule
\end{tabular}

    \caption{Detailed estimates of the standard deviations due to model-related $\sigma$ and bootstrap-related $\tau$ uncertainty in machine translation experiments.}
    \label{tab:varcomp_table_flores_full}
\end{table*}

\begin{table*}[p!]
    \centering
    \resizebox{0.93\textwidth}{!}{
        \begin{tabular}{lrrrrrrrrrrrr}
\toprule
 & \multicolumn{4}{c}{Glot500} 
 & \multicolumn{4}{c}{mBERT}  
 & \multicolumn{4}{c}{XLM-R} \\

 \cmidrule(lr){2-5} \cmidrule(lr){6-9} \cmidrule(lr){10-13}
 
 & \multicolumn{2}{c}{Concat} 
 & \multicolumn{2}{c}{Individual} 
 & \multicolumn{2}{c}{Concat} 
 & \multicolumn{2}{c}{Individual} 
 & \multicolumn{2}{c}{Concat} 
 & \multicolumn{2}{c}{Individual}  \\
 
\cmidrule(lr){2-3} \cmidrule(lr){4-5} \cmidrule(lr){6-7} \cmidrule(lr){8-9} \cmidrule(lr){10-11} \cmidrule(lr){12-13}
 & $\sigma$ & $\tau$ & $\sigma$ & $\tau$ & $\sigma$ & $\tau$ & $\sigma$ & $\tau$ & $\sigma$ & $\tau$ & $\sigma$ & $\tau$  \\
\midrule
$\text{ara}_{\text{AQMAR}}$ & 0.44 & 2.63 & 0.71 & 2.21 & 1.81 & 4.37 & 0.62 & 3.04 & 0.83 & 2.97 & 1.54 & 2.50 \\
$\text{bam}_{\text{MasakhaNER 2.0}}$ & 0.64 & 1.34 & 2.22 & 1.41 & 1.10 & 1.85 & 0.74 & 1.62 & 0.90 & 1.48 & 0.68 & 1.75 \\
$\text{bar}_{\text{BarNER}}$ & 2.52 & 7.97 & 7.43 & 10.91 & 4.57 & 13.64 & 2.34 & 13.96 & 1.89 & 7.94 & 4.27 & 10.80 \\
$\text{bbj}_{\text{MasakhaNER 2.0}}$ & 0.66 & 2.46 & 1.16 & 2.73 & 1.54 & 2.43 & 1.45 & 2.38 & 1.03 & 2.66 & 0.86 & 3.13 \\
$\text{cat}_{\text{AnCora}}$ & 0.06 & 0.53 & 0.50 & 0.53 & 0.18 & 0.98 & 0.16 & 0.66 & 0.24 & 0.52 & 0.14 & 0.45 \\
$\text{cmn}_{\text{UNER Chinese GSDSIMP}}$ & 0.70 & 2.45 & 0.41 & 2.60 & 1.13 & 9.52 & 0.27 & 2.33 & 0.50 & 2.47 & 0.40 & 2.66 \\
$\text{cmn}_{\text{UNER Chinese GSD}}$ & 0.73 & 1.99 & 72.48 & 2.45 & 1.25 & 7.35 & 0.22 & 2.04 & 0.63 & 2.11 & 0.28 & 2.62 \\
$\text{dan}_{\text{DaNE}}$ & 1.83 & 4.02 & 2.67 & 4.01 & 2.53 & 5.61 & 1.12 & 3.96 & 1.49 & 4.10 & 0.49 & 2.96 \\
$\text{deu}_{\text{GermEval}}$ & 0.04 & 0.30 & 0.04 & 0.23 & 0.16 & 0.46 & 0.10 & 0.27 & 0.06 & 0.28 & 0.09 & 0.26 \\
$\text{ell}_{\text{elNER}}$ & 0.10 & 0.46 & 0.06 & 0.39 & 0.22 & 0.79 & 0.15 & 0.52 & 0.08 & 0.41 & 0.02 & 0.43 \\
$\text{eng}_{\text{Tweebank}}$ & 2.66 & 3.11 & 61.11 & 3.53 & 0.68 & 3.97 & 3.46 & 4.13 & 2.66 & 3.23 & 0.74 & 2.81 \\
$\text{eng}_{\text{UNER English EWT}}$ & 0.87 & 1.42 & 0.10 & 1.31 & 0.81 & 1.52 & 0.40 & 1.62 & 0.54 & 1.39 & 0.51 & 1.29 \\
$\text{eus}_{\text{EIEC}}$ & 0.66 & 2.24 & 1.00 & 2.09 & 0.57 & 3.75 & 0.75 & 2.88 & 0.78 & 2.01 & 0.83 & 2.19 \\
$\text{ewe}_{\text{MasakhaNER 2.0}}$ & 0.19 & 0.35 & 0.24 & 0.45 & 0.33 & 0.49 & 0.19 & 0.63 & 0.10 & 0.37 & 0.23 & 0.50 \\
$\text{fin}_{\text{TurkuNLP}}$ & 0.68 & 1.62 & 1.35 & 1.82 & 0.57 & 2.58 & 0.73 & 2.08 & 0.39 & 1.89 & 0.80 & 1.53 \\
$\text{fon}_{\text{MasakhaNER 2.0}}$ & 0.54 & 1.39 & 0.54 & 1.70 & 0.88 & 1.93 & 1.21 & 1.90 & 0.62 & 1.68 & 0.24 & 1.68 \\
$\text{glg}_{\text{SLI Galician Corpora}}$ & 0.58 & 1.04 & 0.34 & 1.18 & 0.69 & 1.40 & 0.43 & 1.32 & 0.64 & 1.00 & 0.19 & 1.13 \\
$\text{hau}_{\text{MasakhaNER 2.0}}$ & 0.26 & 0.42 & 0.52 & 0.45 & 0.46 & 0.74 & 1.34 & 0.95 & 0.23 & 0.43 & 0.42 & 0.44 \\
$\text{hau}_{\text{MasakhaNER}}$ & 0.07 & 1.17 & 0.21 & 1.04 & 0.93 & 1.61 & 0.31 & 1.50 & 0.29 & 1.17 & 0.40 & 0.84 \\
$\text{heb}_{\text{NEMO SPMRL}}$ & 0.33 & 2.03 & 1.79 & 3.48 & 0.68 & 4.80 & 3.01 & 3.10 & 0.60 & 2.05 & 0.51 & 2.84 \\
$\text{heb}_{\text{NEMO UD}}$ & 1.45 & 4.60 & 2.55 & 4.69 & 2.86 & 7.36 & 2.30 & 4.51 & 2.73 & 4.42 & 0.90 & 3.78 \\
$\text{hin}_{\text{HiNER}}$ & 0.09 & 0.02 & 4.34 & 0.01 & 0.01 & 0.04 & 0.00 & 0.02 & 0.01 & 0.02 & 0.00 & 0.01 \\
$\text{hrv}_{\text{hr500k}}$ & 0.11 & 0.32 & 0.08 & 0.28 & 0.17 & 0.55 & 0.15 & 0.47 & 0.06 & 0.32 & 0.17 & 0.32 \\
$\text{ibo}_{\text{MasakhaNER 2.0}}$ & 0.04 & 0.19 & 0.92 & 0.27 & 0.39 & 0.40 & 0.35 & 0.62 & 0.05 & 0.22 & 0.81 & 0.31 \\
$\text{ibo}_{\text{MasakhaNER}}$ & 0.45 & 0.78 & 2.13 & 0.91 & 0.63 & 1.06 & 0.66 & 1.68 & 0.34 & 0.87 & 0.99 & 0.96 \\
$\text{kaz}_{\text{KazNERD}}$ & 0.07 & 0.38 & 6.27 & 0.21 & 0.36 & 1.14 & 0.18 & 0.25 & 0.15 & 0.36 & 790.74 & 0.18 \\
$\text{kin}_{\text{MasakhaNER 2.0}}$ & 0.09 & 0.33 & 0.25 & 0.45 & 0.11 & 0.52 & 0.27 & 0.81 & 0.07 & 0.42 & 0.55 & 0.51 \\
$\text{kin}_{\text{MasakhaNER}}$ & 0.58 & 1.73 & 0.90 & 2.49 & 1.23 & 2.46 & 4.51 & 2.86 & 0.37 & 1.88 & 1.71 & 2.20 \\
$\text{lug}_{\text{MasakhaNER 2.0}}$ & 0.10 & 0.32 & 0.06 & 0.33 & 0.25 & 0.44 & 0.44 & 0.56 & 0.20 & 0.32 & 0.10 & 0.42 \\
$\text{lug}_{\text{MasakhaNER}}$ & 0.46 & 3.47 & 3.27 & 4.14 & 0.51 & 3.97 & 1.58 & 3.32 & 0.48 & 3.38 & 1.82 & 3.92 \\
$\text{luo}_{\text{MasakhaNER 2.0}}$ & 0.27 & 0.70 & 0.39 & 0.72 & 0.11 & 0.90 & 0.34 & 0.94 & 0.20 & 0.72 & 0.40 & 0.85 \\
$\text{luo}_{\text{MasakhaNER}}$ & 0.68 & 4.71 & 187.53 & 7.88 & 1.72 & 5.05 & 4.92 & 8.33 & 1.23 & 4.10 & 4.90 & 5.14 \\
$\text{mar}_{\text{L3Cube MahaNER}}$ & 0.19 & 1.13 & 0.38 & 1.19 & 0.51 & 1.82 & 0.33 & 1.35 & 0.08 & 1.28 & 0.25 & 1.18 \\
$\text{mos}_{\text{MasakhaNER 2.0}}$ & 0.83 & 2.00 & 1.91 & 2.06 & 0.62 & 2.83 & 1.77 & 2.35 & 0.33 & 1.86 & 1.72 & 2.16 \\
$\text{nep}_{\text{EverestNER}}$ & 0.07 & 0.29 & 0.09 & 0.32 & 0.19 & 1.07 & 0.15 & 0.53 & 0.09 & 0.31 & 0.14 & 0.34 \\
$\text{nld}_{\text{CONLL02}}$ & 0.08 & 0.33 & 0.19 & 0.38 & 0.95 & 1.07 & 0.16 & 0.51 & 0.04 & 0.37 & 0.14 & 0.35 \\
$\text{nno}_{\text{NorNE}}$ & 0.15 & 1.31 & 0.30 & 1.28 & 1.91 & 2.74 & 0.52 & 2.03 & 0.22 & 1.33 & 0.70 & 1.29 \\
$\text{nob}_{\text{NorNE}}$ & 0.14 & 0.92 & 0.77 & 1.06 & 2.95 & 2.33 & 1.01 & 1.73 & 0.30 & 0.86 & 0.55 & 0.99 \\
$\text{nya}_{\text{MasakhaNER 2.0}}$ & 0.16 & 0.22 & 0.07 & 0.21 & 0.25 & 0.36 & 0.47 & 0.42 & 0.05 & 0.26 & 0.12 & 0.29 \\
$\text{pcm}_{\text{MasakhaNER 2.0}}$ & 0.15 & 0.45 & 0.30 & 0.49 & 0.45 & 0.79 & 0.65 & 0.67 & 0.11 & 0.52 & 0.09 & 0.54 \\
$\text{pcm}_{\text{MasakhaNER}}$ & 0.14 & 0.96 & 0.75 & 1.33 & 0.50 & 1.59 & 1.03 & 2.30 & 0.44 & 1.25 & 3.29 & 1.77 \\
$\text{por}_{\text{UNER Portuguese}}$ & 0.29 & 0.90 & 0.50 & 0.88 & 0.28 & 1.50 & 0.77 & 1.17 & 0.12 & 0.88 & 0.40 & 0.94 \\
$\text{qaf}_{\text{UNER Arabizi}}$ & 4.37 & 9.44 & 27.94 & 18.19 & 1.55 & 8.86 & 1.49 & 6.32 & 6.06 & 7.12 & 1.01 & 9.57 \\
$\text{ron}_{\text{RONEC}}$ & 0.09 & 0.27 & 0.13 & 0.25 & 0.17 & 0.44 & 0.09 & 0.29 & 0.04 & 0.23 & 0.04 & 0.22 \\
$\text{slk}_{\text{UNER Slovak SNK}}$ & 0.56 & 1.67 & 2.92 & 2.71 & 0.94 & 3.07 & 3.14 & 2.86 & 0.50 & 1.61 & 1.57 & 2.46 \\
$\text{slk}_{\text{WikiGoldSK}}$ & 0.14 & 0.54 & 0.18 & 0.64 & 0.34 & 1.35 & 0.37 & 0.80 & 0.32 & 0.49 & 0.22 & 0.42 \\
$\text{slv}_{\text{ssj500k}}$ & 0.82 & 22.96 & 1.37 & 28.19 & 0.99 & 28.37 & 3.54 & 27.08 & 0.86 & 24.95 & 2.84 & 27.24 \\
$\text{sna}_{\text{MasakhaNER 2.0}}$ & 0.08 & 0.16 & 0.06 & 0.19 & 0.17 & 0.30 & 0.22 & 0.46 & 0.10 & 0.16 & 0.22 & 0.41 \\
$\text{spa}_{\text{AnCora}}$ & 0.11 & 0.29 & 0.03 & 0.22 & 0.12 & 0.42 & 0.07 & 0.31 & 0.05 & 0.28 & 0.05 & 0.22 \\
$\text{spa}_{\text{CONLL02}}$ & 0.06 & 0.47 & 0.21 & 0.46 & 0.09 & 0.50 & 0.04 & 0.49 & 0.28 & 0.46 & 0.16 & 0.40 \\
$\text{swa}_{\text{MasakhaNER 2.0}}$ & 0.04 & 0.11 & 0.04 & 0.12 & 0.06 & 0.20 & 0.14 & 0.20 & 0.02 & 0.12 & 0.05 & 0.14 \\
$\text{swa}_{\text{MasakhaNER}}$ & 0.20 & 0.83 & 0.27 & 1.20 & 0.23 & 1.26 & 0.54 & 1.51 & 0.26 & 0.84 & 0.26 & 1.12 \\
$\text{swe}_{\text{UNER Swedish Talkbanken}}$ & 2.10 & 5.94 & 102.29 & 9.49 & 5.23 & 9.95 & 414.24 & 10.89 & 1.40 & 5.85 & 3.88 & 6.05 \\
$\text{tha}_{\text{ThaiNNER}}$ & 0.07 & 0.31 & 0.06 & 0.27 & 0.08 & 0.40 & 0.06 & 0.48 & 0.06 & 0.31 & 0.06 & 0.27 \\
$\text{tsn}_{\text{MasakhaNER 2.0}}$ & 0.35 & 0.64 & 0.65 & 0.76 & 0.50 & 0.93 & 1.74 & 1.21 & 0.48 & 0.76 & 0.78 & 0.88 \\
$\text{twi}_{\text{MasakhaNER 2.0}}$ & 0.40 & 1.84 & 0.38 & 2.30 & 0.93 & 3.04 & 0.57 & 3.29 & 0.19 & 2.18 & 1.39 & 2.52 \\
$\text{wol}_{\text{MasakhaNER 2.0}}$ & 0.09 & 0.69 & 0.39 & 0.90 & 0.31 & 1.06 & 0.40 & 1.11 & 0.20 & 0.99 & 1.39 & 1.13 \\
$\text{wol}_{\text{MasakhaNER}}$ & 0.73 & 6.40 & 17.24 & 7.39 & 1.60 & 7.20 & 1.65 & 8.22 & 1.31 & 6.19 & 2.26 & 7.67 \\
$\text{xho}_{\text{MasakhaNER 2.0}}$ & 0.12 & 0.37 & 0.11 & 0.43 & 0.20 & 0.71 & 0.20 & 0.90 & 0.17 & 0.39 & 0.44 & 0.46 \\
$\text{yor}_{\text{MasakhaNER 2.0}}$ & 0.20 & 0.63 & 0.83 & 0.63 & 0.65 & 1.05 & 0.35 & 0.78 & 0.51 & 0.66 & 0.32 & 0.68 \\
$\text{yor}_{\text{MasakhaNER}}$ & 0.21 & 2.65 & 8.92 & 2.65 & 0.27 & 2.80 & 0.38 & 2.21 & 0.48 & 2.94 & 1.29 & 2.61 \\
\bottomrule
\end{tabular}

    }
    \caption{Detailed estimates of the standard deviations due to model-related $\sigma$ and bootstrap-related $\tau$ uncertainty in NER experiments.}
    \label{tab:varcomp_table}
\end{table*}

\subsection{Ranks}

Table~\ref{tab:ranktable} shows the marginal distributions of ranks for the NER task.

\end{document}